\documentclass[authoryear,preprint,review,12pt,times]{elsarticle}

\usepackage{amssymb}
\usepackage{lineno}
\usepackage{url}
\usepackage{epsfig}
\usepackage{graphicx}
\usepackage{amsmath}
\usepackage{amssymb}
\usepackage{booktabs}
\usepackage{multirow}
\usepackage{subfigure}
\usepackage{enumitem}
\usepackage[margin=2.5cm]{geometry}
\usepackage{float}
\usepackage{arydshln}

\usepackage{algorithm}
\usepackage{algorithmicx}
\usepackage{algpseudocode}
\usepackage{microtype}

\begin{document}

\begin{frontmatter}



\title{SparseGF: A Height-Aware Sparse Segmentation Framework with Context Compression for Robust Ground Filtering Across Urban to Natural Scenes}

\author[inst1,inst2]{Nannan Qin}
\author[inst3]{Pengjie Tao}
\author[inst1,inst2]{Haiyan Guan}
\author[inst4]{Zhizhong Kang}
\author[inst5]{Lingfei Ma}
\author[inst3]{Xiangyun Hu}
\author[inst6,inst7]{Jonathan Li}

\affiliation[inst1]{organization={School of Remote Sensing and Geomatics Engineering, Nanjing University of Information Science and Technology},%Department and Organization
            % addressline={Address One}, 
            city={Nanjing},
            postcode={210044}, 
            state={JiangSu},
            country={China}}  

\affiliation[inst2]{organization={Tiandu-Nuist Deep Space Exploration Laboratory},%Department and Organization
            % addressline={Address One}, 
            city={Nanjing},
            postcode={210044}, 
            state={JiangSu},
            country={China}}              

\affiliation[inst3]{organization={School of Remote Sensing and Information Engineering, Wuhan University},%Department and Organization
            % addressline={Address One}, 
            city={Wuhan},
            postcode={430079}, 
            state={Hubei},
            country={China}}    

\affiliation[inst4]{organization={School of Land Science and Technology, China University of Geosciences},%Department and Organization
            % addressline={Address One}, 
            city={Beijing},
            postcode={100083}, 
           % state={SiChuan},
            country={China}}   

\affiliation[inst5]{organization={School of Geospatial Artificial Intelligence, East China Normal University},%Department and Organization
            % addressline={Address One}, 
            city={Shanghai},
            postcode={200241}, 
           % state={SiChuan},
            country={China}}       

\affiliation[inst6]{organization={Department of Geography and Environmental Management, University of Waterloo},%Department and Organization
            % addressline={Address Two}, 
            city={Waterloo},
            postcode={N2L 3G1}, 
            state={Ontario},
            country={Canada}}   

\affiliation[inst7]{organization={Department of Systems Design Engineering, University of Waterloo},%Department and Organization
            % addressline={Address Two}, 
            city={Waterloo},
            postcode={N2L 3G1}, 
            state={Ontario},
            country={Canada}}                     

\begin{abstract}
High‑quality digital terrain models derived from airborne laser scanning (ALS) data are essential for a wide range of geospatial analyses, and their generation typically relies on robust ground filtering (GF) to separate point clouds across diverse landscapes into ground and non‑ground parts. Although current deep‑learning‑based GF methods have demonstrated impressive performance, especially in specific challenging terrains, their cross‑scene generalization remains limited by two persistent issues: the context–detail dilemma in large‑scale processing due to limited computational resources, and the random misclassification of tall objects arising from classification‑only optimization. To overcome these limitations, we propose SparseGF, a height‑aware sparse segmentation framework enhanced with context compression. It is built upon three key innovations: (1) a convex-mirror-inspired context compression module that condenses expansive contexts into compact representations while preserving central details; (2) a hybrid sparse voxel-point network architecture that effectively interprets compressed representations while mitigating compression‑induced geometric distortion; and (3) a height‑aware loss function that explicitly enforces topographic elevation priors during training to suppress random misclassification of tall objects. Extensive evaluations on two large‑scale ALS benchmark datasets demonstrate that SparseGF delivers robust GF across urban to natural terrains, achieving leading performance in complex urban scenes, competitive results on mixed terrains, and moderate yet non‑catastrophic accuracy in densely forested steep areas. This work offers new insights into deep‑learning‑based GF research and encourages further exploration toward truly cross‑scene generalization for large‑scale environmental monitoring.
\end{abstract}

%  \centering
%  \includegraphics[width=\linewidth]{Figures/SparseGF.png}

% \begin{highlights}
% \item A convex-mirror-inspired module to resolve the context‑detail dilemma
% \item A hybrid sparse voxel-point network that mitigates geometric distortion
% \item A height‑aware loss penalizing tall‑object misclassification
% \end{highlights}


\end{frontmatter}

%\begin{linenumbers}
\section{Introduction}
\label{sec:intro}
High‑quality digital terrain models (DTMs) are foundational data products for environmental monitoring applications across diverse landscapes, serving as the primary geometric reference for flood inundation analysis in urban floodplains~\citep{Tan2024r, Mihu2025Int}, ecological monitoring in vegetated regions~\citep{mccarley2020estimating, XIANG2024Auto}, and geomorphological hazard assessment in steep natural terrains~\citep{RAZAK2011186, CHEN2014291}. Airborne laser scanning (ALS) has become a primary data source for generating such DTMs, owing to its ability to penetrate vegetation canopies and efficiently collect high‑precision point clouds over large areas. A common fundamental step in this process is ground filtering (GF), which separates raw ALS point clouds into ground and non‑ground parts. Therefore, the ability to perform robust GF from complex urban scenes to natural terrains is of great importance for ensuring the reliability of large-scale environmental monitoring.

Nevertheless, GF continues to face persistent technical challenges due to the complex composition of the Earth's surface. This complexity manifests in two primary ways. First,  significant topographic fluctuations and intricate object structures lead to pronounced intra-class variability. Second, ground and non-ground categories often exhibit remarkably similar geometric characteristics in transitional areas, resulting in high inter-class similarity. Furthermore, geometrically distorted DTMs are frequently generated owing to either the omission of crucial ground points (e.g., mountain ridges or valleys) or the residual presence of low-lying object points (e.g., vegetation on slopes).

To tackle the challenges above, numerous algorithms have been developed over the last few decades. Pioneer GF approaches are knowledge-driven, primarily relying on geometric priors and explicit rule‑based reasoning to achieve robust performance when the underlying assumptions hold. According to different principal concepts, they can be broadly categorized into six groups: (1) morphology-based filters~\citep{Kilian1996capture, zhang2003progressive, pingel2013improved}, (2) surface-based filters~\citep{Pfeifer1999Interpolation,axelsson2000generation, zhang2016easy}, (3) slope-based filters~\citep{vosselman2000slope, sithole2001filtering, susaki2012adaptive}, (4) segmentation-based filters~\citep{sithole_filtering_2005, Hingee2016DIGITAL, Beumier2016Digital}, (5) statistic-based filters~\citep{bartels2006dtm, ozcan2016lidar}, and (6) hybrid filters~\citep{Su2015hierarchical, ZHAO2016Improved, Bigdeli2018DTM}. These knowledge-driven methods typically exhibit high stability and reliability in specific scenarios, but their dependence on explicit geometric assumptions often leads to significant performance degradation in complex environments.

To enhance robustness and automation, researchers have developed more advanced data-driven GF pipelines. Early data-driven approaches~\citep{lu2009hybrid,jahromi2011novel} relied on conventional classifiers to extract discriminative patterns from hand-crafted features, marking a notable departure from traditional knowledge-driven methods. While these methods demonstrated improved adaptability in complex scenes, their performance remained fundamentally limited by the quality and expressiveness of hand-crafted features. Overcoming these limitations, the advent of deep learning (DL) has since revolutionized GF by enabling automatic feature learning, moving beyond hand-crafted features toward end-to-end data-driven paradigms. Depending on the different data formats, existing DL-based GF methods can be categorized into three groups: (1) 2D-image-based pipelines~\citep{hu2016deep, rizaldy2018fully, gevaert2018deep}, (2) 3D-voxel-based pipelines~\citep{Yotsumata2020Qua, Dai2023deep}, and (3) 3D-point-based pipelines~\citep{jin2020point, qin2021opengf, Fareed2023Analysis, Li2025p}. By learning optimal feature representations directly from data with more advanced network architectures, DL-based GF models have achieved increasingly impressive classification accuracy, outperforming prior methods, especially in highly challenging terrains such as steep, forested mountains. However, GF is not merely a binary classification task aimed at maximizing accuracy; rather, it is a comprehensive challenge that involves adapting to diverse terrains, modeling large-scale context, and preserving topographic coherence of the ground surface. 

Consequently, the primary challenge in DL-based GF research has gradually shifted from extracting more powerful discriminative features for ground/non-ground separation to achieving robust performance across diverse terrain scenes. Yet, in striving toward this goal, existing DL-based GF models are hindered by two persistent issues, as highlighted by recent comprehensive GF studies~\citep{QIN2023tow,qin2023deep}:

\begin{enumerate}
\item \textbf{The context–detail dilemma.} The massive data volume of large-scale point clouds forces block-wise processing on conventional hardware, inevitably creating a critical dilemma between contextual continuity and granular detail. The use of smaller blocks allows for the preservation of local details but lacks sufficient context, hindering the identification of large objects. Conversely, larger blocks offer extensive contextual information at the expense of spatial resolution, which impairs the extraction of fine-grained features.
\item \textbf{Random misclassification of tall objects.} Existing methods primarily focus on optimizing classification accuracy, often overlooking elevation constraints during ground point prediction. This leads to objects of various heights being misclassified as ground in an unconstrained manner. Although misclassified points from tall objects, such as roofs or canopies, may represent only a small proportion of the total point cloud, their large elevation deviations from the true ground surface can introduce severe local distortions in the resulting DTMs.
\end{enumerate}

The context–detail dilemma is also pervasive in related geospatial tasks, particularly in urban point cloud semantic segmentation and DSM‑to‑DTM regression. To address this dilemma, efforts have largely centered on multi‑scale strategies. In semantic segmentation, multi‑scale architectures with multiple branches are commonly used to capture large‑scale context from coarse‑resolution inputs while extracting fine details from high‑resolution ones, followed by feature fusion~\citep{Qin2019Semantic, Dai2024large}. However, such approaches introduce inherent drawbacks, including scale‑branch redundancy and the need for carefully designed fusion mechanisms~\citep{Dai2024large}. Moreover, existing multi‑scale strategies are typically tailored and validated for urban scenes, leaving their effectiveness in natural terrains unverified. In DSM‑to‑DTM regression, a similar multi‑scale strategy is applied at test time, where a single‑branch network processes multiple downsampled versions of the input and the resulting predictions are fused to mitigate the influence of large objects~\citep{amirkolaee2022dtm}. This test‑time multi‑scale inference also incurs additional computational overhead and relies on heuristic fusion rules.

The random misclassification of tall objects has also been investigated in related tasks, particularly in urban point cloud semantic segmentation. For instance, incorporating height-above-ground (HAG) as an auxiliary input has been a common remedy, as relative height provides strong priors for distinguishing buildings from ground~\citep{YOUSEFHUSSIEN2018a, ARIEF2019a}. This approach, however, suffers from a circular dependency because accurate HAG estimation relies on a reliable DTM, which itself depends on accurate ground segmentation results. In practice, coarse DTM approximations are used instead. In complex terrains, errors in the approximated HAG can propagate through the network, inadvertently misleading classification rather than enforcing meaningful elevation priors.

Taken together, although many strategies have been proposed in related geospatial tasks to extend advanced DL models to overcome the two core issues, existing approaches still suffer from scale-branch redundancy, circular dependency, and uncertain adaptability to non-urban scenes. To more effectively address these two persistent issues, we propose SparseGF, a height-aware sparse segmentation framework enhanced with context compression to achieve robust performance across diverse landscapes. Extensive evaluations of SparseGF on two large-scale public ALS datasets demonstrate a clear performance gradient, substantially outperforming state-of-the-art methods in complex urban scenes, delivering competitive results on mixed terrains, and maintaining a moderate yet non‑catastrophic level of accuracy in densely forested steep areas. This graded behavior demonstrates the robust cross‑scene generalization of our framework, which stems from three core innovations, each targeting a particular technical challenge:
\begin{enumerate}
\item To resolve the context–detail dilemma most pronounced in urban scenes, we introduce a convex-mirror-inspired context compression module that condenses expansive contexts into compact representations while preserving central details. This design is particularly suited for GF in flat to gently sloping areas, where initial slopes are small, so compression minimally alters the original slopes, and thus has little impact on elevation contexts.

\item To effectively interpret compressed representations, we design a hybrid sparse voxel-point network architecture. The voxel-based backbone captures large context from the entire compressed patch and focuses prediction on the geometrically intact central points, while a lightweight point-wise refinement module enhances feature discriminability of central points and mitigates compression‑induced distortion of contextual information.

\item To suppress random misclassification of tall objects, we devise a height-aware loss function that explicitly enforces topographic elevation priors during training, preserving surface continuity in the final DTM while overcoming the circular dependency in HAG-as-input approaches. Complemented by the point-wise refinement module, this constraint maintains robustness even in steep terrains where the compression module faces extreme challenges.
\end{enumerate}

The remainder of this paper is structured as follows. Section~\ref{sec:method} details the proposed SparseGF framework. Section~\ref{sec:exp_setup} describes the experimental setup, including datasets, evaluation metrics, and implementation details. Section~\ref{sec:results} then presents comprehensive experimental results on two large-scale public ALS datasets, covering comparisons with state-of-the-art methods, ablation studies, error analysis, model interpretability, cross-dataset generalization, and efficiency evaluation. Section~\ref{sec:discussion} provides an in-depth discussion of the overall performance, limitations, and corresponding future research directions, while Section~\ref{sec:conclusion} concludes the paper with a summary and a brief outlook.

\section{The SparseGF framework}
\label{sec:method}
\begin{figure}[ht]
    \setlength{\abovecaptionskip}{0.cm}
    \setlength{\belowcaptionskip}{-0.cm}
    \centering
    \includegraphics[width=0.9\linewidth]{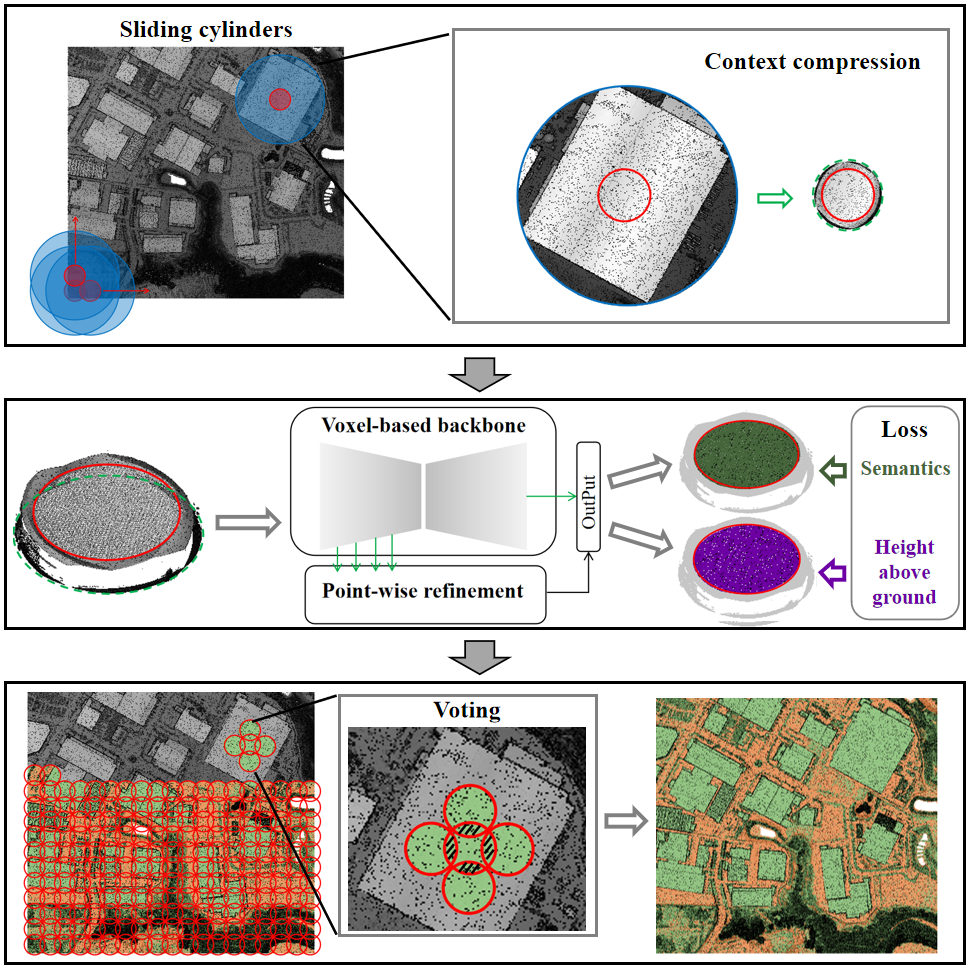}
    \caption{Overview of the proposed SparseGF framework.}
    \label{fig:framework}
\end{figure}

To simultaneously address the two fundamental challenges in DL-based GF, we propose the SparseGF framework that incorporates three complementary core innovations (a context compression module, a hybrid sparse voxel-point network architecture, and a height-aware loss function), enabling robust GF across urban to natural terrains for applications such as floodplain mapping and landslide inventory. The overall workflow of SparseGF, depicted in Fig.~\ref{fig:framework}, proceeds in three key stages.

First, the input point cloud is partitioned into patches by a sliding cylindrical window, a design that ensures directional uniformity for the subsequent compression while boosting neighbor search efficiency. Each patch is then transformed into a compact representation via a convex-mirror-inspired context compression module (Section~\ref{sec:spatialcompression}), which preserves central details while encoding expansive context.

Second, a height-aware sparse segmentation network is trained on the compressed patches (Section~\ref{sec:hybridnetwork}). To prevent the geometric distortion introduced by compression from interfering with feature learning, the network performs classification exclusively on the geometrically intact central region, leveraging the compressed periphery solely as auxiliary context. Additionally, a height-aware loss function is employed during training to explicitly enforce topographic elevation priors, suppressing the random misclassification of tall objects.

Third, during testing, the multiple predictions for points in the overlapping central regions of adjacent compressed patches (i.e., the intersections of the central regions, each indicated by a red circle in Fig.~\ref{fig:framework}) are merged using a soft voting strategy (Section~\ref{sec:merging}).

\subsection{Convex-mirror-inspired context compression module}
\label{sec:spatialcompression}
Large-scale point clouds present significant computational challenges due to their massive data volume, often necessitating block-wise processing on conventional hardware. However, sliding-window partitioning schemes typically fail to capture sufficient semantic context, a limitation especially pronounced in complex urban scenes where small objects (e.g., trees, vehicles) and large structures (e.g., buildings, bridges) coexist. In such environments, effective GF must simultaneously capture fine-grained details and broad spatial context.

Inspired by the way a convex mirror compresses a wide field of view into a compact representation, we propose a context compression module that preserves central details instead, while efficiently condensing peripheral context. Unlike conventional multi-scale architectures that aggregate features from blocks of varying sizes, our approach encodes expansive context into a single compact representation, keeping the central region unchanged. 

\begin{figure}[ht]
    \setlength{\abovecaptionskip}{0.cm}
    \setlength{\belowcaptionskip}{-0.cm}
    \centering
    \includegraphics[width=\linewidth]{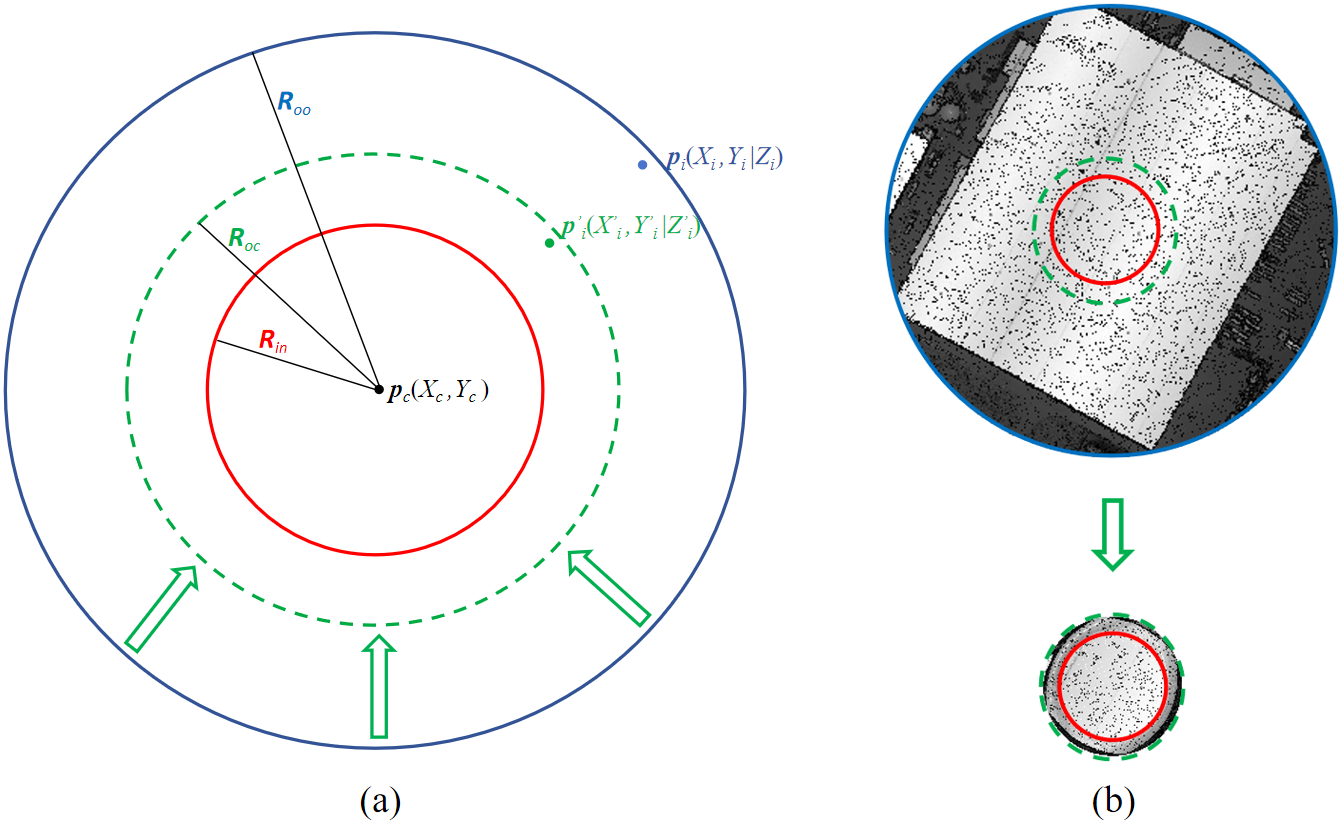}
    \caption{Schematic of the proposed context compression module (2D view): (a) underlying principle and (b) concrete example.}
    \label{fig:sc_p}
\end{figure}

As shown in Fig.~\ref{fig:sc_p}(a), the module operates by defining two key boundaries: an outer original boundary (blue circle) and an inner invariant boundary (red circle). These collectively establish a compression zone between them and an invariant zone inside the inner circle. Planar coordinates within the compression zone are then transformed along the directional arrows (green), effectively mapping the outer original boundary to the outer compressed boundary (green dashed circle). Unlike the optical imaging of convex mirrors, the module does not aim to preserve conformal properties; instead, it is designed as a parametric mapping that prioritizes controllable compression ratios for computational efficiency. 

To mathematically formulate this process, we derive the coordinate mapping from an original point \( P_i(X_i, Y_i, Z_i) \) (projected onto the XY-plane as \( p_i \)) in the compression zone to its compressed counterpart \( P'_i(X'_i, Y'_i, Z'_i) \) (projected as \( p'_i \)), as defined below (see Section~\ref{sec:formula} for the full derivation):

\begin{equation} \label{eq:e8}
 X'_i = \frac{R_{oc} - R_{in}}{R_{oo} - R_{in}} \cdot X_i + \left(1 - \frac{R_{oc} - R_{in}}{R_{oo} - R_{in}}\right) \cdot \left( R_{in} \cdot \frac{X_i - X_c}{\sqrt{(X_i - X_c)^2 + (Y_i - Y_c)^2}} + X_c \right)
\end{equation}

\begin{equation} \label{eq:e9}
 Y'_i = \frac{R_{oc} - R_{in}}{R_{oo} - R_{in}} \cdot Y_i + \left(1 - \frac{R_{oc} - R_{in}}{R_{oo} - R_{in}}\right) \cdot \left( R_{in} \cdot \frac{Y_i - Y_c}{\sqrt{(X_i - X_c)^2 + (Y_i - Y_c)^2}} + Y_c \right)
\end{equation}

\begin{equation} \label{eq:e10}
 Z'_i = Z_i
\end{equation}
where \( R_{oo} \), \( R_{oc} \), and \( R_{in} \) denote the radii of the outer original, outer compressed, and inner invariant boundaries, respectively, while \( (X_c, Y_c) \) denotes the coordinates of the center point \( p_c \).

Based on the derived mapping, a key property of the proposed radial compression is that it preserves elevation information while distorting only horizontal coordinates. Since elevation is unaffected, the distorted periphery remains a reliable source of large-scale elevation cues (i.e., the distribution of surrounding ground and objects in height), which is critical for correctly filtering large objects. 

To further illustrate the module's behavior, we apply it to a cylindrical patch containing a large building, as shown in Fig.~\ref{fig:sc_p}(b). The results demonstrate that the module produces a compact representation that encodes broad spatial context while preserving the central zone, offering a more efficient alternative to multi-scale processing. Although the compressed peripheral region suffers from nonlinear geometric distortion (e.g., a building would exhibit completely different shapes in the compressed periphery versus the central zone), this distortion does not conflict with the spatial invariance assumption of DL models (e.g., standard CNNs) as long as the model only extracts elevation context from the compressed periphery without allowing it to participate in classification. This motivates a dedicated network design, which we present in the following.

\subsection{Height-aware sparse segmentation network}
\label{sec:hybridnetwork}
The proposed network consists of two key components: a hybrid sparse voxel-point network architecture (Section~\ref{sec:architecture}) and a dedicated height-aware loss function (Section~\ref{sec:hagloss}).

\subsubsection{Hybrid sparse voxel-point network architecture}
\label{sec:architecture}
To resolve the conflict between compression-induced geometric distortion and the spatial invariance assumptions of CNNs, we assign distinct roles to the two regions of the resulting compact representation: the compressed periphery serves solely as a source of contextual information, not as a region for classification; consequently, classification and loss computation are confined to the geometrically intact central region, which is crucial for preserving topographic continuity in downstream DTM products used for hydrological modeling or geomorphological hazard assessment.

\begin{figure}[ht]
    \setlength{\abovecaptionskip}{0.cm}
    \setlength{\belowcaptionskip}{-0.cm}
    \centering
    \includegraphics[width=\linewidth]{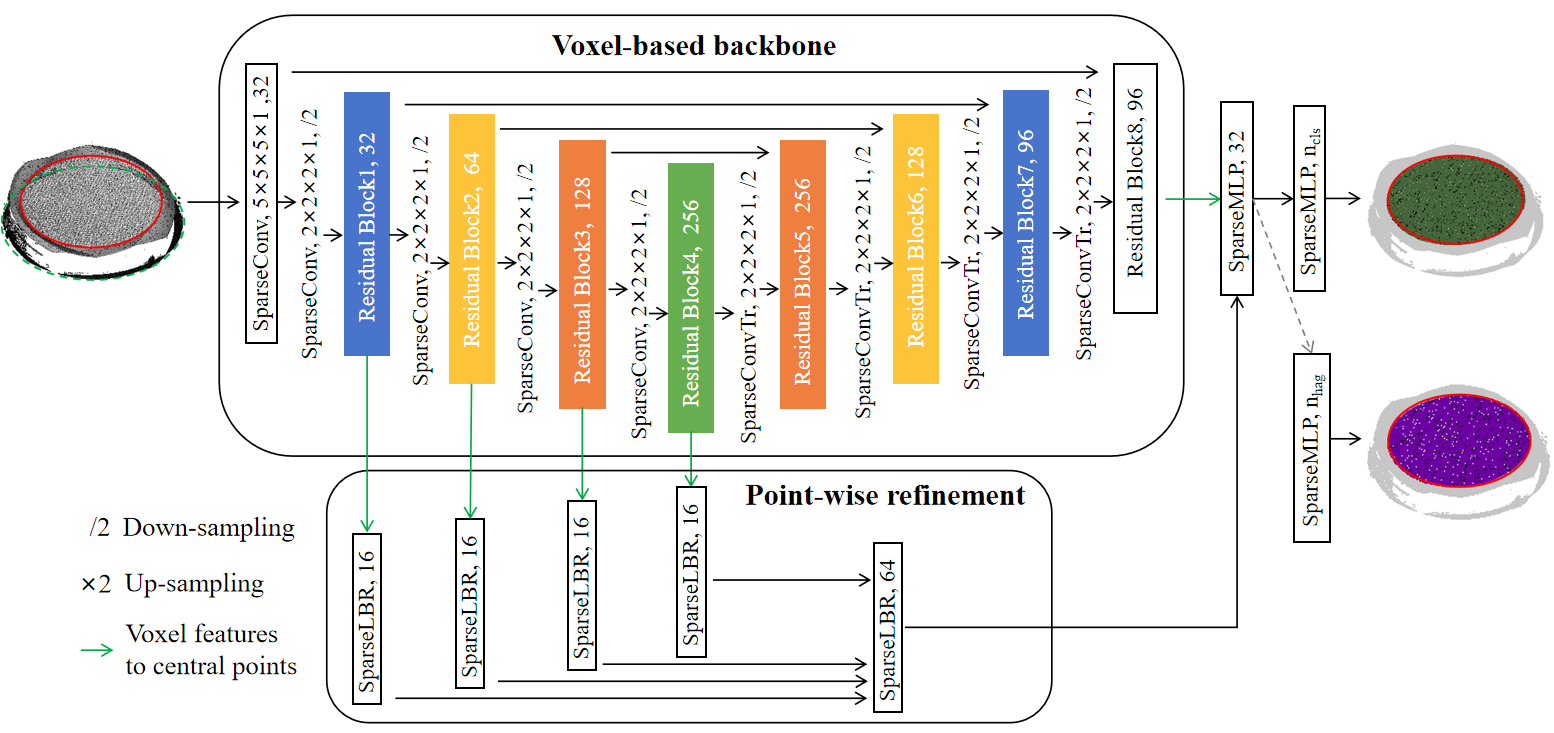}
    \caption{Illustration of the proposed hybrid sparse voxel-point network architecture.}
    \label{fig:archit}
\end{figure}

Guided by this principle, we design a hybrid sparse voxel-point network architecture, illustrated in Fig.~\ref{fig:archit}. It consists of three main modules: a voxel-based backbone, a point-wise refinement module, and multi-task prediction heads. The voxel-based backbone network employs sparse convolutions to process entire compressed point cloud patches efficiently. Intermediate hierarchical voxel features from the backbone encoder are projected back to the central points and fed into the point-wise refinement module for multi-scale point feature enhancement and fusion. Concurrently, the final-layer voxel features from the backbone decoder are also projected to the central points. Both feature streams of the central points are then merged and processed by the multi-task prediction heads for simultaneous semantic classification and HAG estimation.

\textbf{Voxel-based backbone network.} Our architecture employs MinkUNet34C~\citep{choy20194d} as its backbone to process voxelized point cloud patches. This backbone follows a classic U-Net architecture with a symmetric encoder and decoder. The encoder begins with a sparse convolutional layer (kernel size 5, stride 1), followed by four downsampling blocks, each consisting of a sparse convolution (kernel size 2, stride 2) and two cascaded residual blocks. Each residual block contains two sparse convolutional layers (kernel size 3, stride 1) with residual connections. The decoder symmetrically comprises four upsampling blocks, each equipped with a sparse transposed convolution (kernel size 2, stride 2) and two cascaded residual blocks. While this backbone effectively balances efficiency and performance, its voxelized input representation inevitably blurs semantic distinctions by aggregating disparate points into the same voxel. To recover these distinctions, we project the hierarchical voxel features from the four encoder downsampling blocks back to the central points. The projected central point features serve as the foundation for subsequent point-wise feature refinement. Concurrently, we project the final-layer voxel features from the backbone decoder to the central points. Features from the compressed peripheral region are discarded in both projections, as they are excluded from classification and loss computation by design.

\textbf{Point-wise refinement module.} Following the design principle above, this module operates exclusively on the projected hierarchical voxel features of the central points. It consists of five lightweight feed-forward networks (LBRs), each containing a sparse linear layer~\citep{choy20194d}, batch normalization, and ReLU activation.  The sparse linear layer naturally accommodates varying numbers of central points across different blocks, essential because the number of points in the central region changes with each compressed patch, whereas conventional point-wise MLPs would require fixed-size inputs. The module processes the projected hierarchical features in two steps. First, the first four LBRs handle projected features from four different scales, producing multi-scale refined features. These features are then concatenated and fused through the fifth LBR to produce the final enhanced central point features, which are subsequently merged with the final output from the backbone network.

\textbf{Multi-task prediction heads.} The merged feature from the backbone and the refinement module is fed into parallel prediction heads for simultaneous semantic classification and HAG estimation. Each head is implemented as an independent, fully connected layer. The HAG estimation head serves as an auxiliary task that provides topographic elevation priors to regularize the primary classification task. The HAG head is designed to be detached after training, incurring no additional computational overhead during inference. 

\subsubsection{Height-aware loss function}
\label{sec:hagloss}
To suppress misclassification of tall objects (a risk that may be exacerbated by the context compression module in steep areas), we devise a height‑aware loss function that explicitly enforces topographic elevation priors during training.  This is particularly important for applications such as flood modeling and landslide susceptibility analysis, where even a small number of tall-object misclassifications can introduce large elevation errors in DTMs and bias subsequent geomorphic measurements. Unlike classification‑only approaches that use coarse HAG features as input, our loss‑based formulation breaks the circular dependency by treating HAG as a supervisory signal rather than a network input.

While HAG is intrinsically a continuous variable, we formulate its estimation as a classification task by discretizing its values into predefined bins. This approach offers two key benefits: (1) it improves robustness and numerical stability across a wide range of HAG values compared to direct regression, and (2) it simplifies optimization by enabling seamless integration with the primary ground/non-ground classification task within a unified network. Under this formulation, the bin indices increase monotonically with height, so that ground points predominantly fall into lower bins, while tall objects (e.g., building roofs and tree canopies) occupy higher bins.

Building on this monotonic bin structure, we employ a weighted cross-entropy loss. The category-wise weights \( w_c \) increase monotonically with height, reflecting the intuition that misclassifying a taller object as ground introduces larger elevation errors in the DTM and should therefore incur a stronger penalty. The HAG loss \( L_{\text{hag}} \) is defined as:
\begin{equation}
 L_{\text{hag}} = -\frac{1}{N} \sum_{i=1}^{N} \sum_{c=1}^{C} w_c \cdot y_{i,c} \log(\hat{y}_{i,c})
\end{equation}
where \( N \) is the number of points, \( C \) is the number of HAG bins, \( y_{i,c} \) indicates the ground-truth label, \( \hat{y}_{i,c} \) is the predicted probability, and \( w_c \) is the category-wise weight.

To jointly optimize primary ground/non-ground classification and elevation-aware regularization, we combine \( L_{\text{hag}} \) with the standard cross-entropy loss for the classification task \( L_{\text{cls}} \) using a balancing weight \( \lambda \in [0,1] \) and train the entire model end‑to‑end by minimizing the total loss:

\begin{equation} \label{eq:total_loss}
L_{\text{total}} = (1 - \lambda) L_{\text{cls}} + \lambda L_{\text{hag}}
\end{equation}

\subsection{Soft voting for merging predictions from overlapping central regions}
\label{sec:merging}
Testing large-scale ALS point clouds typically requires dividing the scene into smaller blocks due to memory and computational constraints. In our framework, since classification is performed on the circular central region of each compressed patch, covering the entire point cloud necessitates overlapping central regions between adjacent patches. As a result, points in these overlapping areas receive multiple predictions from different patches. Following common practice in point cloud processing, we adopt a simple yet effective soft voting strategy (see Algorithm~\ref{alg:block_merging}) to aggregate the multiple predictions and produce a robust and consistent semantic labeling for the entire point cloud.

\begin{algorithm}[htbp]
\caption{Soft voting for merging predictions from overlapping central regions}
\label{alg:block_merging}
\begin{algorithmic}[1]
\Require The set of all central points from compressed patches, each with a predicted ground probability
\Ensure Consolidated point cloud with globally consistent ground/non-ground labels

\State \textbf{Initialize:} Cumulative probability map $\mathcal{D}$, Vote count map $\mathcal{C}$

\For{each point $p$ with its associated ground probability $prob(p)$}
\State $\mathcal{D}[p] \gets \mathcal{D}[p] + prob(p)$ \Comment{Accumulate probability}
\State $\mathcal{C}[p] \gets \mathcal{C}[p] + 1$ \Comment{Count predictions}
\EndFor

\For{each point $p$ with $\mathcal{C}[p] > 0$}
\State $\bar{prob}_p \gets \mathcal{D}[p] / \mathcal{C}[p]$ \Comment{Compute average probability}
\State $\mathcal{L}[p] \gets \text{\textbf{ground} if } \bar{prob}_p > 0.5 \text{ else \textbf{non-ground}}$
\EndFor

\State \Return Labeled point cloud $\mathcal{L}$
\end{algorithmic}
\end{algorithm}

In summary, the three core innovations of SparseGF work synergistically to address the context–detail dilemma, prevent geometric distortion, and suppress tall‑object misclassification. This collaborative design ensures robust GF across diverse landscapes, thereby supporting reliable DTM generation for large-scale environmental monitoring.
   
\section{Experimental setup}
\label{sec:exp_setup}
In this section, we describe the datasets (Section~\ref{sec:datasets}), evaluation metrics (Section~\ref{sec:metrics}), and implementation details (Section~\ref{sec:implementation}) used to evaluate the robustness of the proposed framework across urban to natural landscapes.

\subsection{Datasets}
\label{sec:datasets}
We adopt two publicly available ALS point cloud datasets for experimental evaluation. Specifically, the OpenGF dataset~\citep{qin2021opengf} is used for performance benchmarking, while an adapted version of the DFC2019 dataset~\citep{Saux2019Gr} is used for evaluating cross-dataset generalization. 

\textbf{OpenGF dataset.} OpenGF is a recently published large-scale GF benchmark. It encompasses diverse terrain scenes collected from four different countries, covering a total area of approximately 47.7~$km^2$ and containing over 542 million points. A total of 160 standardized point cloud tiles (each 500 × 500~$m^2$) are used for model training and validation. The test set includes three point clouds, each with distinct characteristics. Test I covers multiple typical terrain types with moderate slopes and vegetation density, representing mixed landscapes. Test II features complex urban scenes with substantial size variations of buildings and structures, representing complex urban environments. Test III comprises densely forested mountainous regions characterized by extremely sparse ground points and sharply changing elevations on terraced slopes, representing challenging natural terrains. These three test samples form a clear gradient from urban to natural terrains, enabling a systematic evaluation of GF methods across diverse landscapes. It is worth noting that this study employs the outlier-excluded version of Test II.

\textbf{Adapted DFC2019 dataset.} To evaluate the cross-dataset generalization of our method, we adapted the DFC2019 dataset for GF tasks. DFC2019 was originally designed for semantic segmentation and is known for its diversity of large objects. Our adapted version comprises four 512×512~$m^2$ test samples that span a range of terrain types: S1 (large buildings), S2 (dense bridges and buildings), S3 (steep slopes with dense vegetation), and S4 (large buildings with dense vegetation). This selection covers scenarios from complex urban structures to natural terrains. The original annotations comprised five categories: ground, building, tree, water, and bridge deck. For GF, all non‑ground classes were merged into a single category. Subsequently, the dataset was manually inspected and corrected using CloudCompare (v2.13.2, \url{https://www.cloudcompare.org}) to ensure label accuracy. 

\begin{figure}[ht]
    \setlength{\abovecaptionskip}{0.cm}
    \setlength{\belowcaptionskip}{-0.cm}
    \centering
    \includegraphics[width=\linewidth]{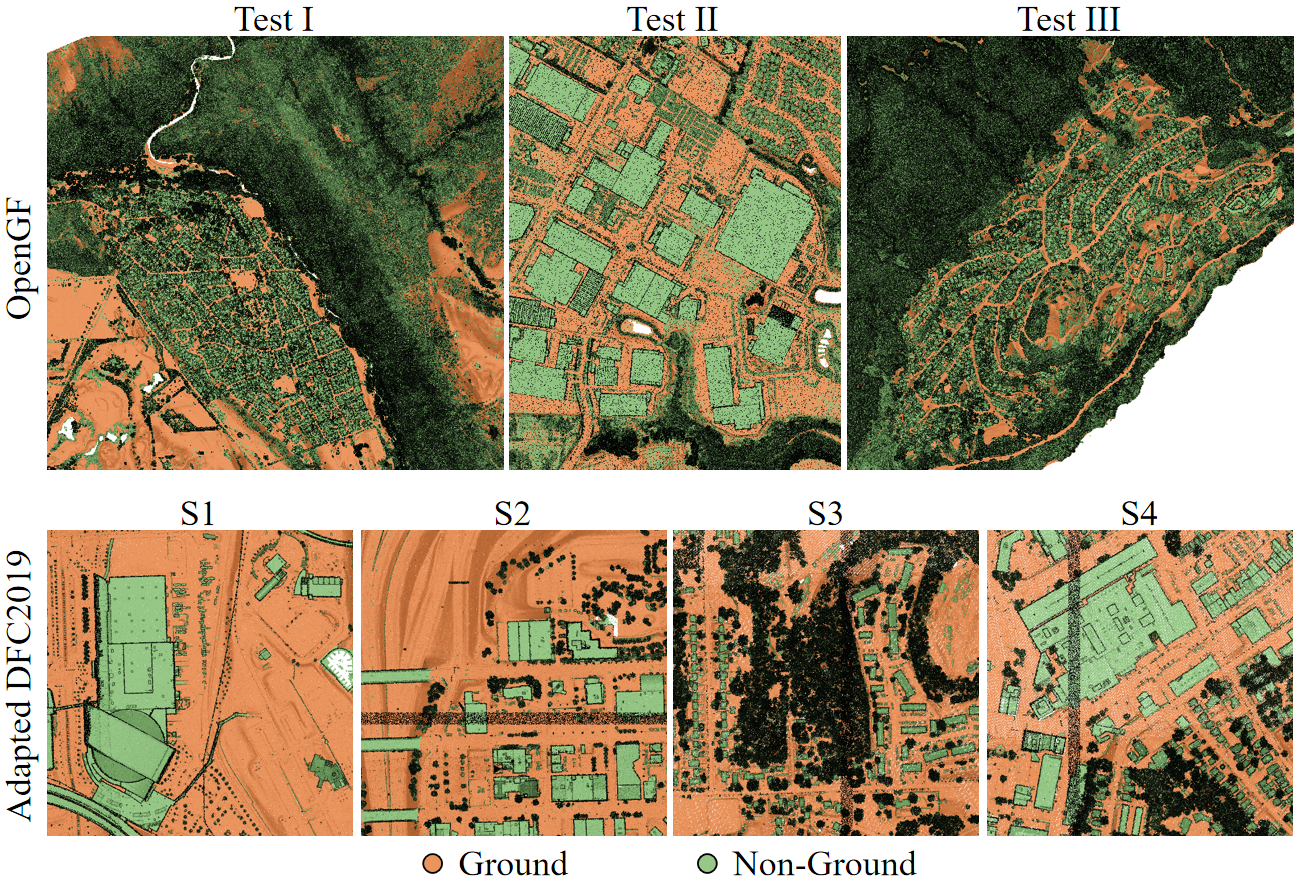}
    \caption{Visualization of the test samples from the OpenGF and adapted DFC2019 datasets.}
    \label{fig:visTest}
\end{figure}

Fig.~\ref{fig:visTest} visualizes all test samples from both datasets, and Table~\ref{tab:statistic} summarizes their detailed characteristics.

% \usepackage{multirow}
\begin{table}[H]
\centering
\caption{Statistics of the test samples from the OpenGF and adapted DFC2019 datasets.}
\begin{tabular}{cccccccccc}
\hline
\multirow{2}{*}{} & \multicolumn{3}{c}{OpenGF}  & \multicolumn{6}{c}{Adapted DFC2019} \\ \cline{2-10} 
                  & Test I & Test II & Test III  & S1 & S2  & S3 & S4 \\ \cline{1-10} 
Area~($km^2$)            & 6.60    &  1.10       &   2.30        &  0.26   &  0.26     &  0.26   &   0.26  \\
Points~($M$)          & 46.80     &  6.22       &     19.80       &   1.52 &  1.61   &  1.29  &  0.82 \\
Density~($points/m^2$)         & 7.09   &   5.65      &   8.61         & 5.85   &  6.19   &  4.96  &  3.15  \\\hline
\end{tabular}
\label{tab:statistic}
\end{table}

\subsection{Evaluation metrics}
\label{sec:metrics}
Following the OpenGF benchmark~\citep{qin2021opengf}, we employ three metrics for evaluation: per-class intersection-over-union ($IoU$), overall accuracy ($OA$), and root mean square error ($RMSE$) between the predicted and reference DTMs. These metrics are formally defined as follows:

\begin{equation}  \label{eq:iou}
IoU_1 = \frac {TP_1}{TP_1 + FP_1 + FP_2}, \quad 
IoU_2 = \frac {TP_2}{TP_2 + FP_2 + FP_1}
\end{equation}

\begin{equation}  \label{eq:oa}
OA =  \frac {TP_1 + TP_2}{TP_1 + TP_2 + FP_1 + FP_2}, \quad 
RMSE = \sqrt{\frac{1}{N} \sum_{i=1}^{N} (P_i - R_i)^2}
\end{equation}
where \(TP_1\)/\(FP_1\) denotes the number of non-ground points correctly/incorrectly classified, while \(TP_2\)/\(FP_2\) are defined analogously for ground points. \(N\) refers to the number of valid pixels in the DTM, while \(P_i\) and \(R_i\) represent the elevation values of the \(i\)-th valid pixel in the predicted and reference DTMs, respectively.

\subsection{Implementation details}
\label{sec:implementation}

\textbf{Data partitioning and compression.} We partition the input point cloud using a sliding cylindrical window with a radius of 150~m and a step size of 50~m. The context compression module is configured with an outer compressed radius \(R_{oc} = 44\)~m and an inner invariant radius \(R_{in} = 25\sqrt{2}\)~m (i.e., \(\sqrt{2}/2\) times the step size). The voxel size for sparse convolution is set to 0.5~m. To clearly demonstrate the contribution of each proposed component, we also evaluate the backbone network as a standalone baseline. Due to the computational burden of directly processing the original 150~m-radius cylindrical patches, this baseline is limited to a voxel size of 1.0~m. During compression, we record an index mask that identifies which points belong to the central region; this mask is later used to project features to the central points and confine loss computation.

\textbf{Height-aware loss settings.} Continuous HAG values (in m) are quantized into six discrete bins based on typical object heights: ground \(\{0\}\), low grass \((0, 0.2]\), low shrubs \((0.2, 0.5]\), medium shrubs \((0.5, 1.0]\), low structures or small trees \((1.0, 3.0]\), and tall structures or large trees \((3.0, +\infty)\). For the weighted cross‑entropy loss \(L_{\text{hag}}\), the category weights \(w_c\) are set to increase linearly from 1 to 6 across the six bins. In the total loss \(L_{\text{total}}\), the balancing coefficient \(\lambda\) is set to 0.5.

\textbf{Implementation language.} The hybrid sparse voxel-point network is implemented in Python using PyTorch. The preprocessing (partitioning and compression) and soft voting merging are implemented in C++ for efficiency. Currently, the preprocessing is performed in a single thread. Since each patch is processed independently, the preprocessing time can be further reduced using hardware acceleration (e.g., multi‑threading).

\textbf{Training details.} During training, outliers are explicitly treated as non‑ground points by relabeling their original class from 0 to 1. The model is optimized using the Adam optimizer with a batch size of 4. The initial learning rate is set to \(1 \times 10^{-1}\), and a cosine annealing scheduler is applied to reduce it to \(1 \times 10^{-3}\) over 12 epochs. Training incorporates 3D data augmentation through random rotation around the vertical (z-) axis. All experiments were conducted on a workstation equipped with an NVIDIA GeForce RTX 3090 GPU, an Intel Xeon W‑2245 CPU (3.90 GHz), and 64 GB of RAM.

We also provide a detailed sensitivity analysis of key hyperparameters (e.g., the balancing coefficient \(\lambda\), the number of HAG bins, and the outer compressed radius \(R_{oc}\)) in Section~\ref{sec:hyperparameter}.

\section{Results and analysis}
\label{sec:results}
This section presents the experimental results. We first evaluate the overall performance of our method on the OpenGF benchmark, including quantitative comparisons with state-of-the-art methods, qualitative comparisons with the backbone baseline, and an analysis of challenging local areas. Next, ablation studies isolate the contributions of each core component, followed by error analysis and model interpretability. We then assess cross-dataset generalization on the adapted DFC2019 dataset and finally analyze computational efficiency.

\subsection{Benchmark overall performance}
\label{sec:opengfresult}
\subsubsection{Quantitative comparison with state-of-the-art methods}
Table~\ref{tab:resOpengf} summarizes a quantitative comparison of SparseGF with state-of-the-art methods on the OpenGF benchmark, along with its backbone baseline as a reference. The compared methods include PTD~\citep{axelsson2000generation}, PMF~\citep{zhang2003progressive}, MCC~\citep{Evans2007multiscale}, CSF~\citep{zhang2016easy}, PointNet++~\citep{qi2017pointnet++}, KPConv~\citep{thomas2019kpconv}, RandLA-Net~\citep{hu2020randla}, SCF-Net~\citep{fan2021scf}, SeqGP~\citep{Dai2023deep}, and PointTransformer~\citep{Xu2026E}.

% \usepackage{multirow}
% \usepackage{graphicx}
\begin{table}[ht]
\centering
\caption{Quantitative comparison of different GF methods on the OpenGF benchmark.}
\label{tab:resOpengf}
\resizebox{\textwidth}{!}{%
\begin{tabular}{ccccccccccccc}
\hline
\multirow{2}{*}{Method} & \multicolumn{4}{c}{Test I}                     & \multicolumn{4}{c}{Test II}                     & \multicolumn{4}{c}{Test III}                     \\ \cline{2-13} 
                        & $OA$ (\%) & $RMSE$ (m) & $IoU_1$ (\%) & $IoU_2$ (\%)  & $OA$ (\%) & $RMSE$ (m) & $IoU_1$ (\%) & $IoU_2$ (\%)  & $OA$ (\%) & $RMSE$ (m) & $IoU_1$ (\%) & $IoU_2$ (\%) \\ \hline
PTD~\citep{axelsson2000generation}                     & 94.82   & 0.37     & 91.10      & 89.00        & 93.30    & 0.12     & 87.64     & 87.24     & 97.55   & 3.24     & 97.05     & 87.16     \\
PMF~\citep{zhang2003progressive}                     & 90.63   & 0.51     & 85.22     & 79.62     & 86.56   & 1.11     & 78.50      & 73.61     & 94.93   & 0.99     & 94.09     & 73.67     \\
MCC~\citep{Evans2007multiscale}                     & 96.29   & 0.42     & 93.63     & 91.86     & 84.44   & 1.63     & 75.39     & 70.27     & 96.97   & 1.29     & 96.40      & 84.12     \\
CSF~\citep{zhang2016easy}                     & 93.07   & 0.95     & 88.17     & 85.64     & 89.34   & 1.89     & 81.08     & 80.38     & 95.35   & 1.77     & 94.42     & 78.29     \\
PointNet++~\citep{qi2017pointnet++}              & 97.58   & 0.25     & 95.75     & 94.68     & 87.38   & 4.89     & 75.19     & 79.63     & 98.12   & 3.64     & 97.72     & 90.24     \\
KPConv~\citep{thomas2019kpconv}                  & \textbf{97.79}   & 0.20     & \textbf{96.10}     & \textbf{95.17}     & 91.09   & 3.87     & 82.44     & 84.67     & \textbf{98.31}   & \textbf{0.79}     & \textbf{97.94}     & \textbf{91.28}     \\
RandLA-Net~\citep{hu2020randla}              & 96.29   & 0.29     & 93.74     & 91.65     & 94.46   & 1.20      & 90.38     & 90.42     & 97.60    & 1.72     & 97.14     & 87.08     \\
SCF-Net~\citep{fan2021scf}                 & 95.75   & 0.28     & 92.90     & 90.43     & 90.91   & 2.88     & 83.35     & 83.32     & 97.23   & 2.94     & 96.70      & 85.18     \\
SeqGP~\citep{Dai2023deep}                   & 96.52   & 0.23     & 94.12     & 92.15     & 95.20    & 0.32     & 90.90      & 90.78     & --    &    --      &   --        &  --  \\
PointTransformer~\citep{Xu2026E}      & 97.43   & 0.26     & 95.10     & 92.47     & 93.92    & 1.26     & 89.51      & 87.38     & 97.46    &   1.87     &   96.98       &  88.12  \\ \hline
Backbone              & 94.39   & 0.28     &  90.72         &    87.58      & 93.98   & 2.28     &   88.27       &    89.01     & 97.00    & 2.34     &  96.42         &    84.37     \\
SparseGF                   & 96.92   & \textbf{0.19}     & 94.73      & 93.07     & \textbf{95.75}   & \textbf{0.08}     & \textbf{91.75}     & \textbf{91.93}     & 97.86   & 2.26     & 97.43     & 88.53    \\ \hline
\end{tabular}%
}
\end{table}

As shown in Table~\ref{tab:resOpengf}, SparseGF delivers leading performance on Test II (complex urban terrain), substantially surpassing all existing state‑of‑the‑art methods. Prior to SparseGF, SeqGP was the best DL method, and PTD was the best traditional method on this challenging urban site. SparseGF outperforms both by a clear margin, achieving 95.75\% $OA$ (vs. 95.20\% and 93.30\%), 0.08 m $RMSE$ (vs. 0.32 m and 0.12 m), and higher $IoU$ for both ground and non‑ground classes. 

Moving to Test I (mixed terrain), SeqGP and PTD show significant degradation compared to their performance on Test II. They perform markedly worse than the best methods on this terrain, where KPConv was the best DL method, and MCC was the best traditional method. In contrast, SparseGF consistently delivers superior results across all metrics compared to SeqGP and PTD, achieving an $RMSE$ of 0.19 m, comparable to the previous best of 0.20 m (KPConv). Its OA (96.92\%) is slightly lower than KPConv (97.79\%) but higher than MCC (96.29\%).

Finally, on Test III (densely forested, mountainous area), compared with KPConv and MCC, SparseGF achieves an $OA$ of 97.86\%, which is slightly below KPConv’s 98.31\% but above MCC’s 96.97\%. Its $RMSE$ (2.26 m) is higher than both KPConv’s 0.79 m and MCC’s 1.29 m. Nevertheless, it still maintains a moderate yet non‑catastrophic level of accuracy and outperforms PTD (97.86\% $OA$ vs. 97.55\%, 2.26 m $RMSE$ vs. 3.24 m). Notably, while KPConv and MCC achieve better or comparable results on Test III than SparseGF, they are substantially outperformed by SparseGF on Test II. SparseGF outperforms KPConv by 0.67\% in $OA$ and 0.71 m in $RMSE$, and outperforms MCC by 2.31\% in $OA$ and 1.21 m in $RMSE$. 

Overall, these results reveal a clear performance gradient of our method from urban to natural terrains. SparseGF may not be the best on every individual terrain, but it maintains consistently strong performance across diverse landscapes, whereas methods that excel on a specific terrain often fail elsewhere.

This same gradient is also reflected in the terrain‑dependent performance gains of SparseGF over its backbone baseline (Table~\ref{tab:resOpengf}). The improvement is largest on urban Test II, moderate on mixed Test I, and smallest on steep forested Test III. This trend can be explained by the behavior of the context compression module and the presence of large objects in different terrains. In flat urban terrains, the initial slope is small, so compression minimally alters the original slopes, thus having little impact on the elevation contextual cues. Moreover, urban scenes typically contain abundant large objects, and the large‑scale context provided by the compression module is particularly beneficial for their identification. In contrast, on steep natural terrains, compression further increases the steepness of slopes by reducing horizontal distances, and the resulting slope distortion can generate misleading contextual cues for the central region. Furthermore, large objects are scarce in such terrains, so the peripheral context offers little added value. Consequently, the benefit of context compression diminishes as the terrain becomes steeper, which directly explains why the performance gain over the backbone baseline decreases from Test II to Test III.

\begin{figure}[ht]
\centering
\includegraphics[width=\linewidth]{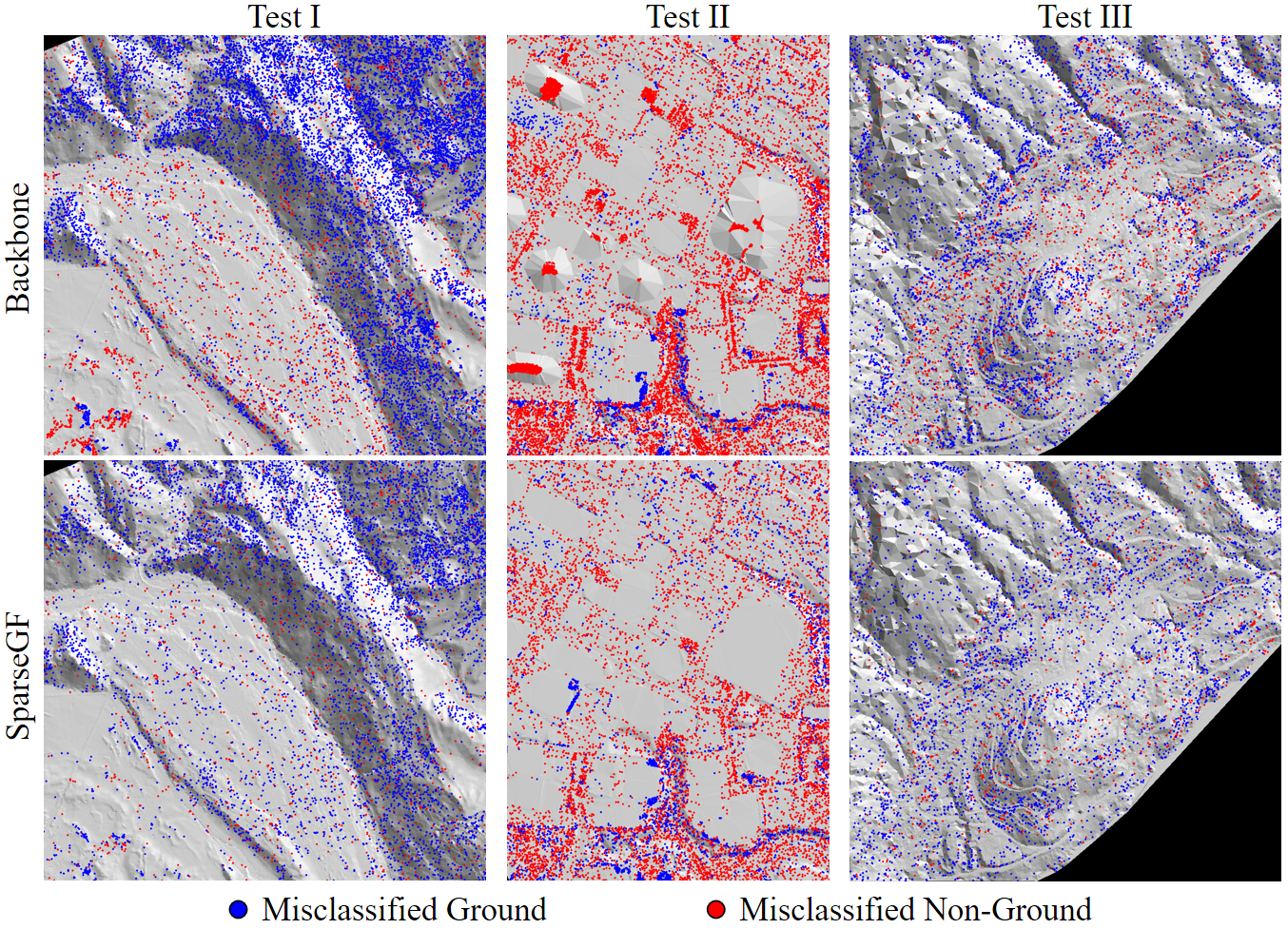}
\caption{Qualitative comparison between SparseGF and its backbone baseline on the OpenGF dataset.}
\label{fig:resOpengf}
\end{figure}

\subsubsection{Qualitative comparison with the backbone baseline}
Complementing the quantitative comparison, Fig.~\ref{fig:resOpengf} presents a qualitative comparison between SparseGF and its backbone baseline, with key errors highlighted to facilitate performance interpretation.

The visual results corroborate the terrain‑dependent behavior observed in the quantitative evaluation. In complex urban regions (Test II), where the backbone baseline struggles most with misclassifying large building roofs, SparseGF substantially reduces such errors, benefiting from the reliable elevation context. On mixed terrains (Test I), SparseGF also achieves noticeable reductions in both ground and non-ground misclassifications, with a particularly pronounced decrease in the misclassification of non‑ground points. This can be attributed to the point‑wise refinement module enhancing point‑wise feature discrimination and the height‑aware loss suppressing the misclassification of tall objects. In steep forested areas (Test III), where compression introduces more severe geometric distortion and large objects are scarce, the improvement of SparseGF over the backbone baseline is becoming weaker.

\subsubsection{Analysis on challenging local areas}
\label{sec:local_analysis}
\begin{figure}[ht]
    \centering
    \includegraphics[width=\linewidth]{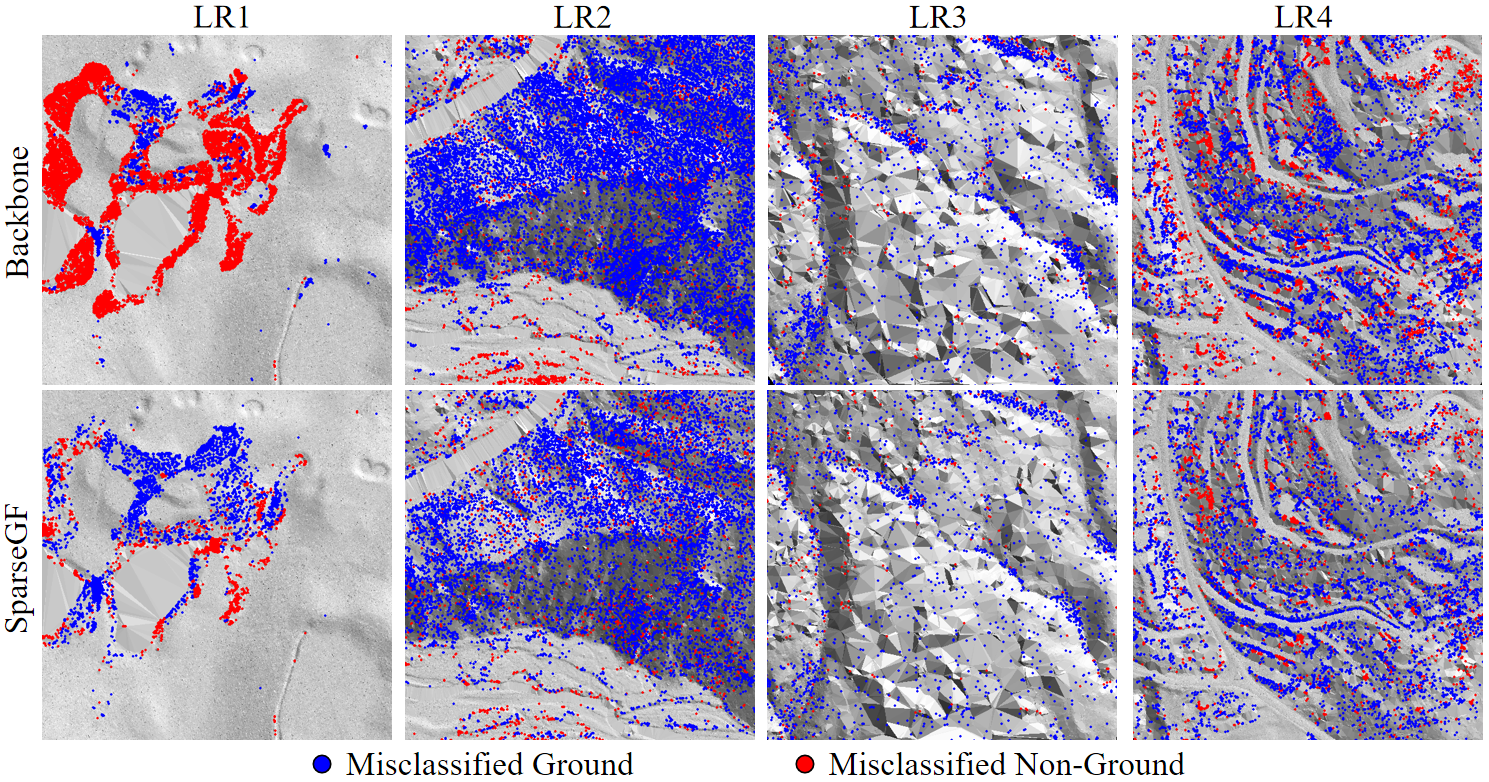}
    \caption{Qualitative comparison between SparseGF and its backbone baseline in four local areas.}
    \label{fig:resLocal}
\end{figure}

Building upon the overall evaluation, we conduct a detailed analysis of several challenging local areas from Tests I and III to further evaluate method performance in complex micro-terrain. Following the previous study~\citep{qin2023deep}, we analyze four representative local regions: two from Test I (LR1 and LR2) and two from Test III (LR3 and LR4). Table~\ref{tab:MicroRes} provides a quantitative comparison between SparseGF and its backbone baseline in these areas, while Fig.~\ref{fig:resLocal} visualizes misclassified points to offer an intuitive depiction of method behavior.

% \usepackage{graphicx}
\begin{table}[ht]
\setlength{\abovecaptionskip}{0.cm}
\setlength{\belowcaptionskip}{-0.cm}
\centering
\caption{Quantitative comparison between SparseGF and its backbone baseline in four local areas.}
\label{tab:MicroRes}
\resizebox{\columnwidth}{!}{%
\begin{tabular}{ccccccccccccccccc}
\midrule
   \multirow{2}{*}{Method} & \multicolumn{4}{c}{$OA$~(\%)}  & \multicolumn{4}{c}{$RMSE$~(m)}  & \multicolumn{4}{c}{$IoU_1$~(\%)}    & \multicolumn{4}{c}{$IoU_2$~(\%)}  \\\cline{2-17}
    & LR1    & LR2    & LR3   & LR4    & LR1    & LR2    & LR3   & LR4    & LR1    & LR2    & LR3   & LR4 & LR1    & LR2    & LR3   & LR4    \\ \midrule
Backbone  & 92.50  & 91.80 & 98.68 & 95.09 & 0.17  & 0.69 & 5.68 & 0.58 & 57.30  & 90.13 & 98.65 & 92.92 & 91.66 & 67.31 & 62.96 & 86.18 \\
SparseGF  & \textbf{97.41} & \textbf{95.06} & \textbf{99.08} & \textbf{96.55} & \textbf{0.13} & \textbf{0.40} & \textbf{5.63} & \textbf{0.39} & \textbf{86.05} &  \textbf{93.82}  & \textbf{99.05}   & \textbf{94.97}  &  \textbf{96.91}    & \textbf{80.22}  & \textbf{73.39}    & \textbf{90.08}    \\\bottomrule
\end{tabular}%
}
\end{table}

As shown in Table~\ref{tab:MicroRes} and Fig.~\ref{fig:resLocal}, the improvement achieved by SparseGF over its backbone baseline is more pronounced in these challenging local areas than on the entirety of Tests I and III (see Table~\ref{tab:resOpengf}). This consistent and enhanced performance gain demonstrates the effectiveness of our proposed innovations in handling complex micro‑terrains.

\begin{figure}[H]
    \centering
    \includegraphics[width=\linewidth]{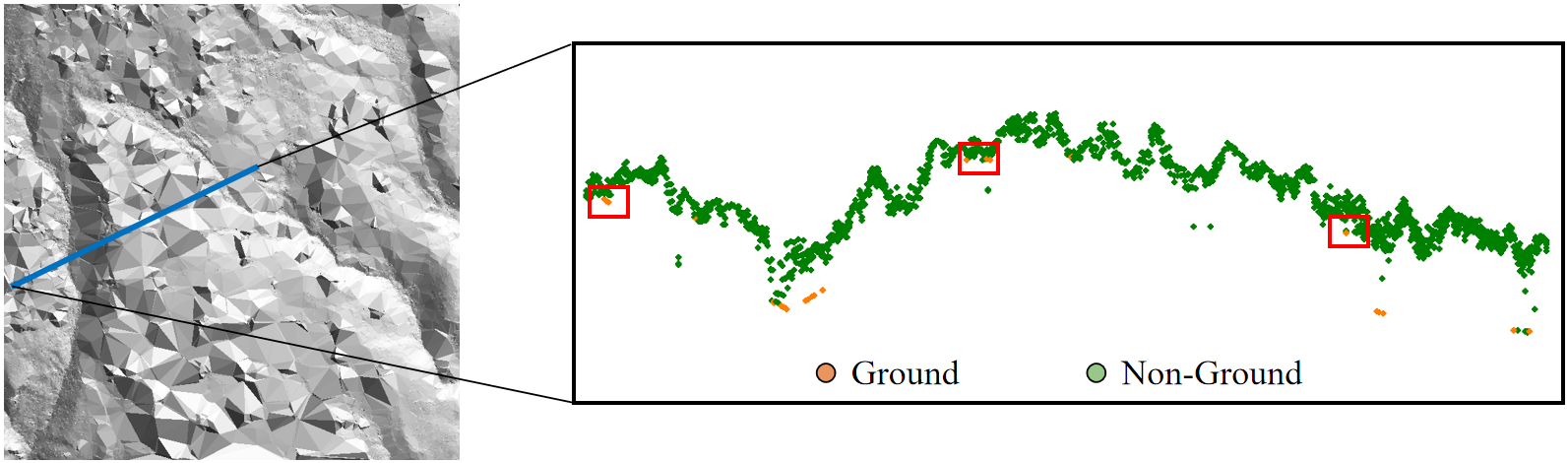}
    \caption{Investigation of pronounced DTM errors through profile analyses.}
    \label{fig:errorDEM}
\end{figure}

However, in the densely forested area LR3, the DTM generated by SparseGF exhibits a pronounced artifact, reflected in a high $RMSE$ of 5.63 m (only marginally better than the backbone’s 5.68 m). To further investigate the origin of this artifact, we extract and analyze a profile from the affected region (see Fig.~\ref{fig:errorDEM}). The analysis reveals that in areas with exceptionally sparse ground points, points from the lowest vegetation layer are erroneously classified as ground (marked by the red rectangle in Fig.~\ref{fig:errorDEM}). Although these misclassifications are few and thus have a limited impact on overall classification accuracy, their elevation is substantially higher than the actual bare ground. This elevation discrepancy directly leads to protruding artifacts in the DTM, which severely degrade its accuracy. A deeper interpretability analysis of this error pattern will be further discussed in the error analysis (Section~\ref{sec:error_analysis}).

\subsection{Ablation studies}
\label{sec:ablation}

\subsubsection{Component effectiveness}
\begin{figure}[ht]
    \centering
    \includegraphics[width=\linewidth]{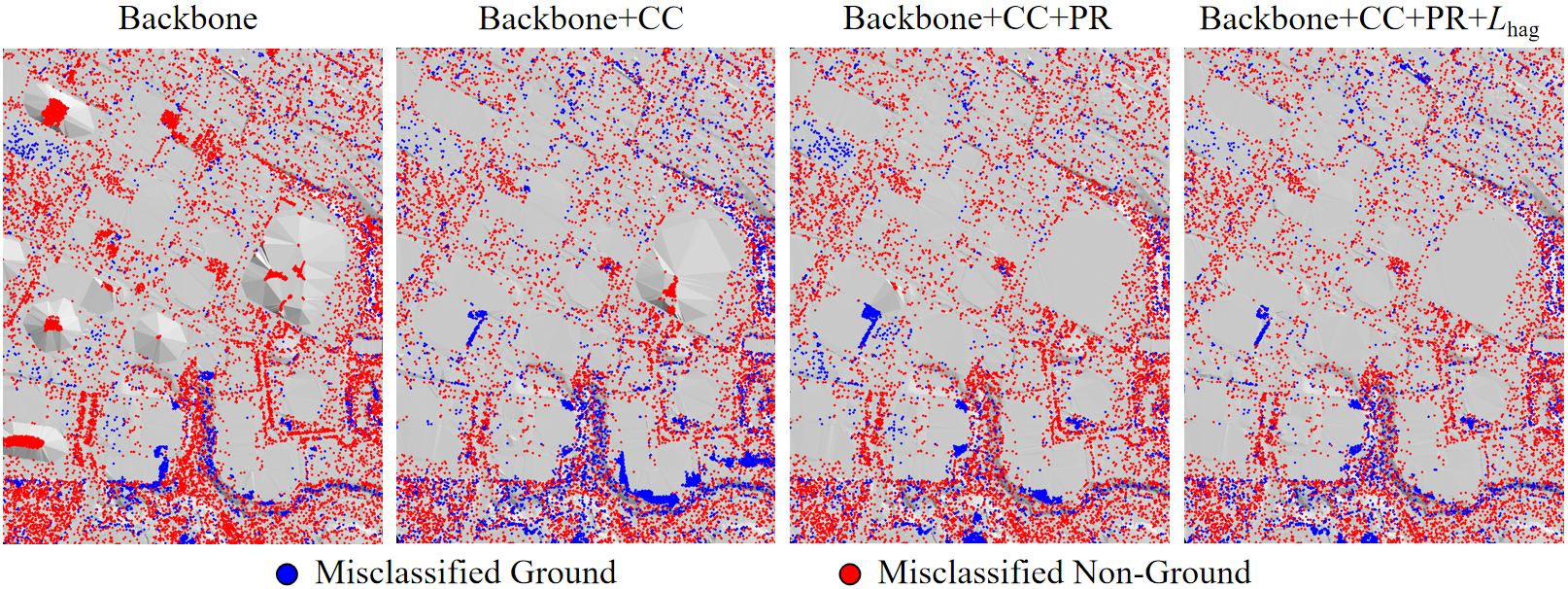}
    \caption{Qualitative ablation results on Test II of the OpenGF dataset. “CC” denotes the context compression module, “PR” signifies the point-wise refinement module.}
    \label{fig:sc}
\end{figure}

To further isolate the contribution of each proposed innovative component, we analyze component effectiveness on the OpenGF dataset. Starting from the backbone baseline, we progressively integrate the context compression module, followed by the individual and combined incorporation of the point-wise refinement module and the HAG loss \( L_{\text{hag}}\). The quantitative results of this progressive integration are summarized in Table~\ref{tab:ablation}, and Fig.~\ref{fig:sc} illustrates the qualitative improvements on Test II.

\begin{table}[ht]
\centering
\caption{Ablation study of the SparseGF framework on the OpenGF dataset. “CC” denotes the context compression module, “PR” signifies the point-wise refinement module.}
\label{tab:ablation}
\begin{tabular}{cccccccccc}
\hline
\multirow{2}{*}{Method} & \multirow{2}{*}{CC} & \multirow{2}{*}{PR} & \multirow{2}{*}{\( L_{\text{hag}}\)} & \multicolumn{3}{c}{$OA$ (\%)} & \multicolumn{3}{c}{$RMSE$ (m)} \\ \cline{5-10} 
& & & & Test I & Test II & Test III & Test I & Test II & Test III \\ \hline
Backbone & & & & 94.39 & 93.98 & 97.00 & 0.28 & 2.28 & 2.34 \\
+CC & \checkmark & & & 96.44 & 95.18 & 97.50 & 0.24 & 1.26 & 3.46 \\
+CC+PR & \checkmark & \checkmark & & 96.89 & 95.62 & \textbf{97.90} & 0.23 & 0.20 & 2.91 \\
+CC+$L_{\text{hag}}$ & \checkmark & & \checkmark & 96.54 & 95.73 & 97.65 & 0.20 & 0.30 & 2.52 \\
SparseGF & \checkmark & \checkmark & \checkmark & \textbf{96.92} & \textbf{95.75} & 97.86 & \textbf{0.19} & \textbf{0.08} & \textbf{2.26} \\ \hline
\end{tabular}
\end{table}

Compared to the backbone baseline, the configuration with only the compression module (+CC in Table~\ref{tab:ablation}) improves $OA$ on all test sites and reduces $RMSE$ on Test II (urban terrain) and Test I (mixed terrain), benefiting from preserved elevation context and higher central resolution (0.5 m vs. 1.0 m). On Test III (steep forested area), however, adding only the compression module leads to increased $RMSE$ due to the slope distortion and misleading contextual cues introduced by compression in such terrains. Nevertheless, the $OA$ still exceeds that of the backbone baseline, partly due to the finer voxel resolution (0.5 m) in the central region. This graded behavior reflects the inherent difficulty of simultaneously achieving context compression and preserving geometric invariance in a single compression module.

Further, the inclusion of either the point-wise refinement module or the HAG loss brings consistent improvements in both $OA$ and $RMSE$ across all test data. Specifically, compared to using only the compression module, adding the point-wise refinement module (the +CC+PR configuration in Table~\ref{tab:ablation}) improves both $OA$ and $RMSE$, as does adding the HAG loss (the +CC+$L_{\text{hag}}$ configuration). This indicates that both enhancing feature extraction (via the point‑wise refinement module) and embedding topographic elevation priors (via the HAG loss) are effective strategies for boosting filtering performance. The complete SparseGF framework achieves superior performance on nearly all metrics, with a notable $RMSE$ of 2.26 m in Test III, substantially recovering from the 3.46 m observed when the compression module was used alone. This reversal demonstrates the efficacy of the point-wise refinement module and HAG loss in jointly compensating for the distortion of contextual information caused by compression, thereby securing the overall performance advance.

\subsubsection{Comparison of HAG loss formulations}
To determine the optimal loss formulation for HAG estimation in SparseGF, we compare the classification‑based \( L_{\text{hag}}\) (weighted cross‑entropy) with a regression variant using Smooth L1 loss~\citep{Girshick2015f}. The comparative results on the OpenGF dataset are summarized in Table~\ref{tab:hag}.

\begin{table}[ht]
\centering
\caption{Comparison of different loss types for HAG estimation on the OpenGF dataset.}
\label{tab:hag}
\begin{tabular}{ccccccc}
\hline
\multirow{2}{*}{Loss type} & \multicolumn{3}{c}{$OA$ (\%)}                  & \multicolumn{3}{c}{$RMSE$ (m)}                       \\ \cline{2-7} 
                       & Test I             & Test II             & Test III            & Test I             & Test II             & Test III            \\ \hline
Regression (Smooth L1)  & 96.68 & 95.47 & 97.78 &    0.21      &    1.37       &    2.35       \\ 
Classification (Weighted cross-entropy)                & \textbf{96.92}          & \textbf{95.75} & \textbf{97.86}          & \textbf{0.19} & \textbf{0.08} & \textbf{2.26} \\\hline
\end{tabular}
\end{table}

As shown in Table~\ref{tab:hag}, the classification‑based formulation consistently outperforms the regression alternative across all test sets, especially on Test II, where $RMSE$ improves from 1.37 m to 0.08 m. This result validates the advantages of the classification‑based design discussed in Section~\ref{sec:hagloss} (improved robustness and numerical stability, as well as seamless integration with the primary classification task).

\begin{figure}[ht]
    \centering
    \includegraphics[width=\linewidth]{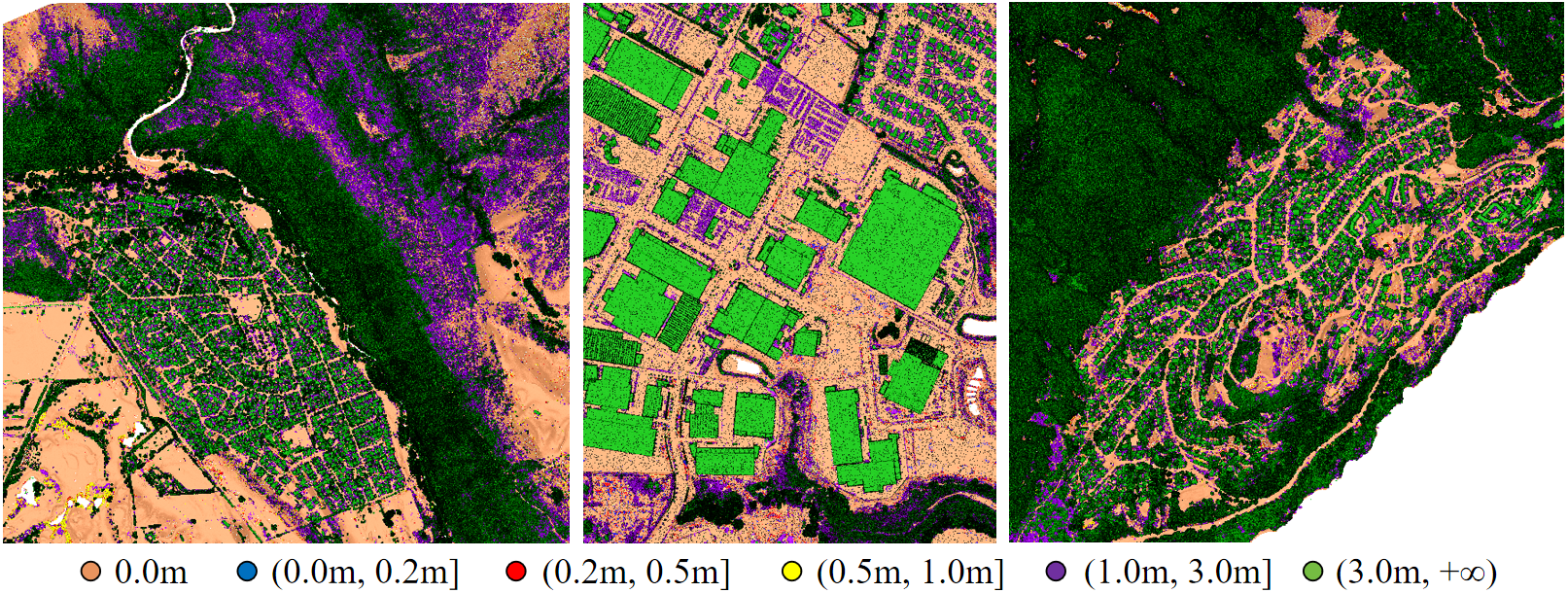}
    \caption{Qualitative HAG estimation results by SparseGF on the OpenGF dataset.}
    \label{fig:hag}
\end{figure} 

We also evaluate the accuracy of the classification‑based HAG estimation as an auxiliary task. Despite its auxiliary role, the HAG prediction achieves high $OA$ scores of 91.54\%, 92.31\%, and 94.11\% on Tests I, II, and III, respectively. The strong quantitative findings are further supported by the qualitative results in Fig.~\ref{fig:hag}, where our method delivers reliable HAG predictions across all test data. Together, they validate that the network successfully learns discriminative features for accurate HAG perception.

\subsubsection{Effect of lovász loss}
While the Lovász loss~\citep{Berman2018The} has demonstrated notable success in semantic segmentation by mitigating class imbalance, its effectiveness in the context of GF remains largely unexamined. To explore its potential, we integrate it as an additional regularization term into the complete SparseGF framework. Quantitative results of the framework without and with the Lovász loss on the OpenGF dataset are presented in Table~\ref{tab:Lovasz}.

% \usepackage{multirow}
\begin{table}[ht]
\centering
\caption{Quantitative comparison of SparseGF with and without Lovász loss on the OpenGF dataset.}
\label{tab:Lovasz}
\begin{tabular}{ccccccc}
\hline
\multirow{2}{*}{Method} & \multicolumn{3}{c}{$OA$ (\%)}                  & \multicolumn{3}{c}{$RMSE$ (m)}                      \\ \cline{2-7} 
                       & Test I             & Test II             & Test III            & Test I             & Test II             & Test III            \\ \hline
SparseGF             & 96.92          & \textbf{95.75} & 97.86          & \textbf{0.19} & \textbf{0.08} & \textbf{2.26} \\
SparseGF + Lovász loss  & \textbf{97.00} & 95.61 & \textbf{97.91} & \textbf{0.19}          & 0.10          & 2.49          \\ \hline
\end{tabular}
\end{table}

As shown in Table~\ref{tab:Lovasz}, adding the Lovász loss yields a slight improvement in $OA$. However, this gain is accompanied by a consistently higher $RMSE$. This trade-off suggests that the Lovász loss, while further enhancing overall classification accuracy, may increase the misclassification of ground points essential for micro‑terrain reconstruction, potentially degrading DTM quality. Therefore, the Lovász loss is not included in our final framework.

\subsection{Error analysis and model interpretability}
\label{sec:error_analysis}

\subsubsection{Error distribution}
\label{sec:error_distribution}
\begin{figure}[ht]
    \centering
    \includegraphics[width=\linewidth]{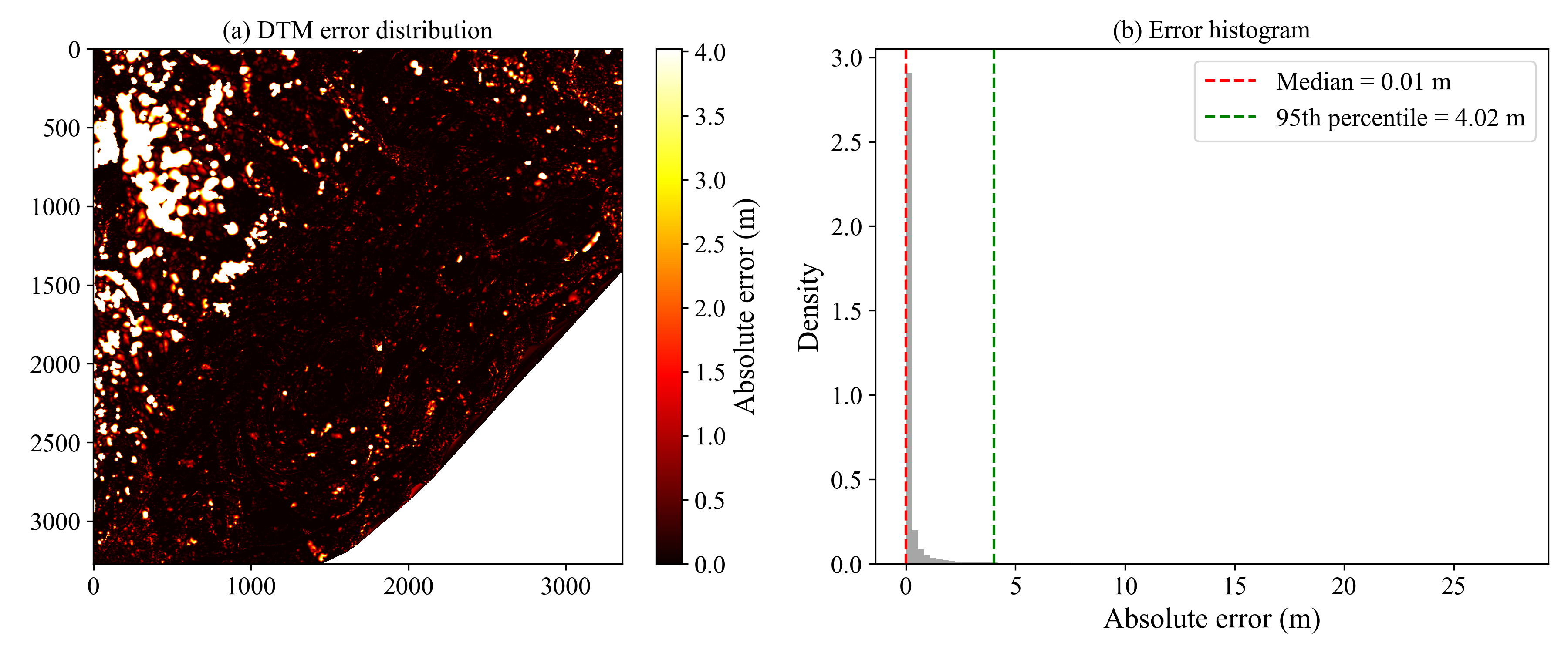}
    \caption{Absolute DTM residuals for SparseGF on Test III: (a) spatial distribution and (b) histogram.}
    \label{fig:error_distribution}
\end{figure}

Given that SparseGF achieves only marginal improvement over its backbone baseline in steep forested areas (Test III) and underperforms several state-of-the-art methods on this site (Table~\ref{tab:resOpengf}), we analyze the error distribution of SparseGF within Test III to identify the spatial patterns and terrain factors associated with its limited effectiveness. The results are presented in Fig.~\ref{fig:error_distribution}.

By comparing the error distribution (Fig.~\ref{fig:error_distribution}(a)) with the original point cloud (Fig.~\ref{fig:visTest}) and the qualitative results (Fig.~\ref{fig:resOpengf}), we observe that errors are not uniformly distributed but concentrate in regions with three characteristic properties, including extremely steep slopes, dense vegetation cover (e.g., coniferous forests with high canopy closure), and complex surface morphology (e.g., ridge–valley transitions). In contrast, sparsely vegetated steep slopes or building areas exhibit considerably lower errors, suggesting that the combination of steep slope and dense vegetation is particularly challenging. The histogram in Fig.~\ref{fig:error_distribution}(b) reveals a median error of 0.01~m and a 95th percentile of 4.02~m. Further computation shows that 90\% of the pixels have errors below 1.0~m, whereas the top 5\% of error regions exhibit a mean error of 9.78~m, underscoring the severe local distortion caused by these challenging terrain features.

To quantify the terrain dependence, we compute the correlation between error magnitude and several geometric attributes. The Pearson correlation coefficient between residual error and local slope angle is 0.17, while the correlation with the standard deviation of normal vectors (a proxy for planarity violation) reaches 0.14. These relatively weak positive correlations indicate that while steepness and non-planarity contribute to errors, other factors, such as vegetation density and data sparsity, also play significant roles. Notably, within the top 5\% error regions, the mean slope reaches 32.0\textdegree{}, confirming that the compression module's effectiveness is most severely degraded on irregular, steep, and densely vegetated terrains.

\subsubsection{Feature importance}
\label{sec:feature_importance}
To understand the contributions of the voxel‑based backbone and the point‑wise refinement module in challenging steep forested areas, we performed a gradient‑based feature importance analysis. We randomly selected 100 compressed patches from Test III and then randomly sampled 1000 points from the geometrically intact central regions of these patches. For each sampled point, we computed the gradient of the classification loss with respect to two feature groups: the voxel‑based backbone features and the point‑wise refinement features. The L2 norm of the gradient reflects the sensitivity of the loss to changes in each feature group, serving as a proxy for their relative importance.

The results show that the point‑wise refinement module accounts for approximately 64.4\% of the total gradient, while the voxel‑based backbone accounts for 35.6\%. This indicates that the point‑wise refinement module plays a dominant role in the final classification decision in steep forested areas. Its effectiveness is further supported by the ablation study (Section~\ref{sec:ablation}), where adding the point‑wise refinement module alone (to the compression module) reduces the $RMSE$ in Test III from 3.46 m to 2.91 m.

The voxel-based backbone extracts features from the entire compressed patch, including the distorted periphery. This distorted context degrades the discriminability of the features corresponding to the central region, leading to a lower gradient contribution. In contrast, the point-wise refinement module operates exclusively on the features projected to the central points, enhancing local discriminability and mitigating distortion-induced effects. Consequently, its features have a stronger direct influence on the classification loss.

\subsubsection{Height-aware loss behavior}
\label{sec:behavior}
The height-aware loss is designed to suppress misclassification of tall objects by enforcing topographic elevation priors. To understand why it effectively suppresses errors in mixed terrain (Test I) but fails in steep forested areas (Test III), we examine its spatial behavior.

On Test I (mixed terrain with moderate slopes and vegetation density), SparseGF achieves the lowest $RMSE$ among all methods, despite its $OA$ not being the highest. This indicates that even when more misclassifications occur, the elevation errors introduced into the DTM are significantly reduced, thanks to the height‑aware loss. The HAG loss imposes progressively larger penalties on higher elevation bins, discouraging the network from misclassifying tall objects (e.g., tree canopies) as ground. This outcome validates the efficacy of the HAG loss in terrains where sufficient ground references exist.

In contrast, Test III presents a fundamentally different challenge: the extreme scarcity of ground points in some densely vegetated areas. The profile analysis (Fig.~\ref{fig:errorDEM}) reveals that low-lying vegetation is frequently misclassified as ground in areas with exceptionally sparse or even absent ground points, indicating that the HAG loss function loses its effectiveness under such conditions. Without reliable ground references, the network cannot adequately distinguish low-lying vegetation from true ground. Consequently, points from the lowest vegetation layer are frequently misclassified as ground. These misclassifications, though few in number, introduce large elevation deviations that significantly increase $RMSE$ while only marginally affecting $OA$. This explains why SparseGF maintains competitive $OA$ in Test III but suffers from increased $RMSE$, reflecting the failure of the HAG constraint under extremely sparse ground points.

\subsection{Cross-dataset generalization}
\label{sec:dfcresult}
To assess the cross-dataset generalization of SparseGF, we directly apply it and its backbone baseline (both pre-trained on the OpenGF dataset) to test samples from the adapted DFC2019 dataset without any fine-tuning. For benchmarking, comparisons are made against the PMF and CSF methods under their respective optimal parameters. The parameter search for PMF covered maximum window sizes (25~m, 50~m, 100~m, 200~m), slopes (0.2, 0.5, 1.0), and cell sizes (0.5~m, 1.0~m, 2.0~m). For CSF, the search included rigidness (2, 3) and cloth resolution (0.5~m, 1.0~m, 2.0~m). In each case, the configuration achieving the highest $OA$ was selected for the final comparison.

\begin{figure}[ht]
    \centering
    \includegraphics[width=\linewidth]{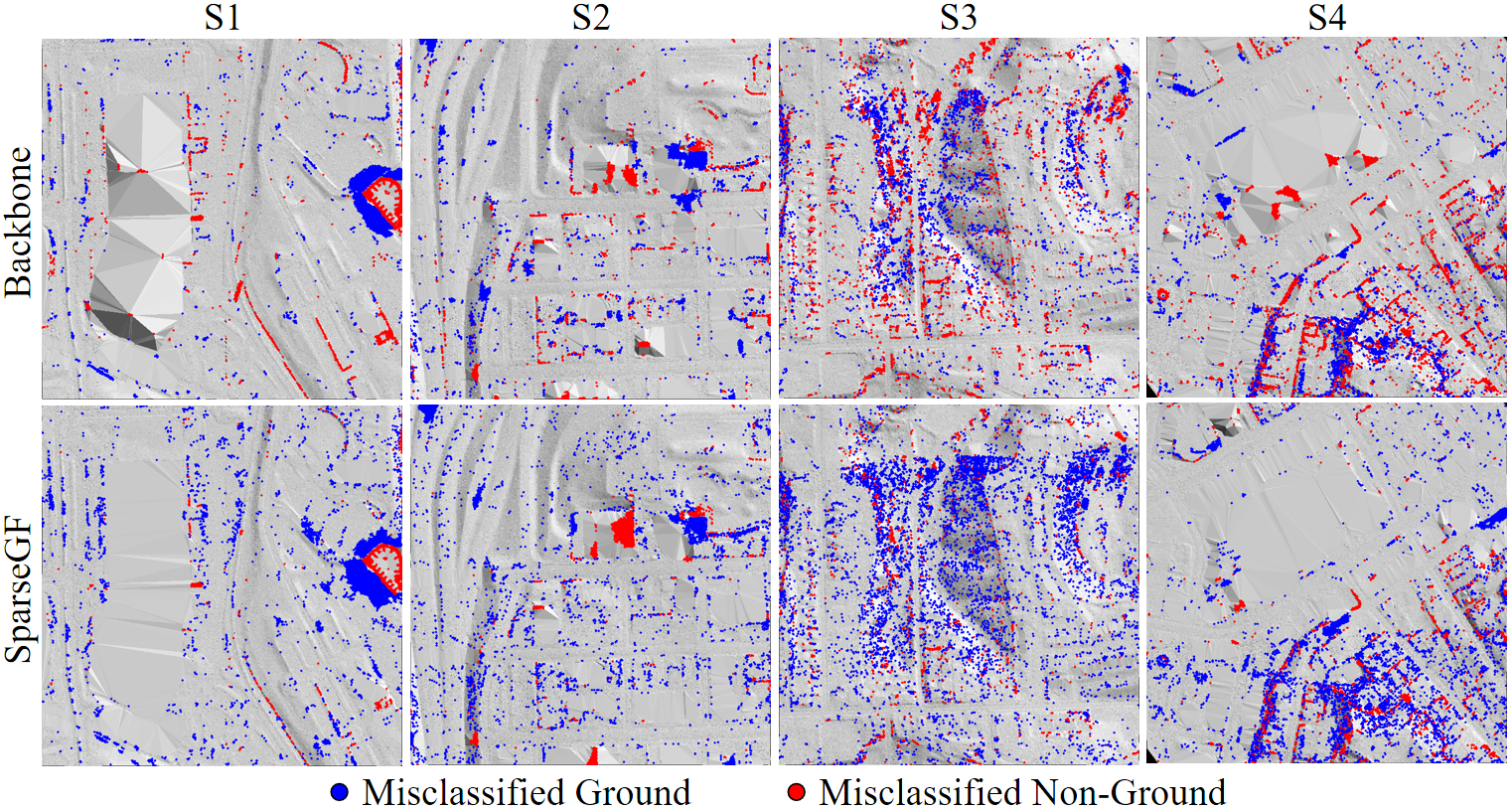}
    \caption{Qualitative comparison between SparseGF and its backbone baseline on the adapted DFC2019 dataset.}
    \label{fig:res_DFC}
\end{figure}

Quantitative comparisons of different GF methods on the adapted DFC2019 dataset are summarized in Table~\ref{tab:res_DFC}, while a qualitative comparison between SparseGF and its backbone baseline is provided in Fig.~\ref{fig:res_DFC}. Both the quantitative and qualitative results confirm that SparseGF has strong cross-dataset generalization capability, substantially outperforming traditional GF methods across all four test sites. 

% \usepackage{multirow}
% \usepackage{graphicx}
\begin{table}[ht]
\centering
\caption{Quantitative comparison of different GF methods on the adapted DFC2019 dataset.}
\label{tab:res_DFC}
\resizebox{\textwidth}{!}{%
\begin{tabular}{ccccccccc}
\hline
\multirow{2}{*}{Method}                                                 
        & \multicolumn{2}{c}{S1}                & \multicolumn{2}{c}{S2}                                  & \multicolumn{2}{c}{S3}                           & \multicolumn{2}{c}{S4}        \\ \cline{2-9} 
      & \multicolumn{1}{l}{$OA$ (\%)} & $RMSE$ (m)          & \multicolumn{1}{l}{$OA$ (\%)} & \multicolumn{1}{l}{$RMSE$ (m)}        & \multicolumn{1}{l}{$OA$ (\%)} & \multicolumn{1}{l}{$RMSE$ (m)} & $OA$ (\%)            & $RMSE$ (m)          \\ \hline
PMF~\citep{zhang2003progressive}     & 78.48    & 10.17 &     74.32     &       0.94    &   79.40   &  0.46    &     81.51    & 1.67 \\ 
CSF~\citep{zhang2016easy}     & 95.37    & 0.65 &    95.19     &      1.71     &   94.37   &  0.50    &     91.45     & 1.56 \\ \hline
Backbone          & \textbf{97.98}         & 4.19          & \textbf{97.98}                  & 0.78                               & \textbf{96.79}                  & 0.11                     & \textbf{96.91} & 0.88          \\
SparseGF      & 97.39             & \textbf{0.55}      & 97.75         & \textbf{0.53}          & 96.00        & \textbf{0.09}    & 96.88         & \textbf{0.78} \\ 
\hline
\end{tabular}%
}
\end{table}

Additionally, SparseGF achieves higher DTM accuracy (lower $RMSE$) than its backbone baseline, with only a marginal decrease in $OA$. Qualitatively, the backbone baseline exhibits frequent misclassification on parts of large buildings, whereas SparseGF delivers more robust performance. These results further confirm that the proposed method maintains robust GF across terrains ranging from urban to natural environments, consistent with the findings on the OpenGF dataset.

\subsection{Efficiency analysis}
Beyond filtering accuracy, we evaluate the practical efficiency of SparseGF by comparing its parameter count and time costs with those of its backbone baseline, as presented in Table~\ref{tab:efficiency}. To ensure a fair comparison, both methods follow the same processing pipeline (preprocessing, model inference, and soft voting merging). Their time differences come solely from the context compression step (applied only in SparseGF) and the differences in network architecture. The soft voting merging step takes the same amount of time for both methods (only 0.75 min on OpenGF and 0.04 min on adapted DFC2019) and is therefore not included in the table.

% \usepackage{multirow}
\begin{table}[ht]
\centering
\caption{Efficiency comparison between SparseGF and its backbone baseline.}
\label{tab:efficiency}
\begin{tabular}{cccccc}
\hline
\multirow{2}{*}{Model} & \multicolumn{2}{c}{Preprocessing time (min)} & \multicolumn{2}{c}{Inference time (min)} & \multirow{2}{*}{Params (M)} \\ \cline{2-5}
                       & OpenGF            & Adapted DFC2019      & OpenGF & Adapted DFC2019 &                         \\ \hline
Backbone         &       \textbf{5.17}            &         \textbf{0.29}           & 8.24      &  0.43       &       \textbf{37.85}               \\
SparseGF               &     5.97            &      0.34                  & \textbf{6.12}       &   \textbf{0.35}      &    37.86        \\ \hline
\end{tabular}
\end{table}

\textbf{Preprocessing time.} The backbone baseline requires only patch partitioning. SparseGF additionally performs radial compression, which increases preprocessing time from 5.17 min to 5.97 min on OpenGF (an increase of about 15\%) and from 0.29 min to 0.34 min on the adapted DFC2019 dataset (an increase of about 17\%). This overhead is expected and is the price for obtaining a compact input representation.

\textbf{Inference time and parameter count.} The context compression module dramatically reduces the number of voxels processed by SparseGF, leading to substantially faster inference despite the extra point‑wise refinement module. On OpenGF, inference time drops from 8.24 min to 6.12 min (a 26\% reduction); on adapted DFC2019, from 0.43 min to 0.35 min (19\% reduction). The refinement module adds only around 0.01 M parameters (from 37.85 M to 37.86 M), which is negligible compared to the voxel reduction benefit.

Overall, despite a modest increase in preprocessing time, SparseGF achieves a substantial reduction in inference time, resulting in lower overall processing time (by 1.32 min on OpenGF and 0.03 min on the adapted DFC2019 dataset).

\section{Discussion and limitations}
\label{sec:discussion}
\subsection{Overall performance across diverse terrains}
The proposed SparseGF demonstrates terrain-adaptive robustness across diverse landscapes, delivering robust GF across urban to natural terrains. It achieves substantial improvements over state-of-the-art methods in complex urban scenes where large objects and micro-terrains intertwine, delivers competitive performance on mixed terrains (on par with the best existing methods), and maintains a moderate yet non‑catastrophic level of accuracy in densely forested steep mountainous areas.

This graded performance reflects an inherent trade‑off between context compression and geometric invariance. In urban scenes (e.g., OpenGF Test II), where both large-scale context and fine details are essential, the compression module excels by providing both through the compressed large-scale elevation context and the preserved central region. In mixed terrains (e.g., OpenGF Test I), the benefits remain substantial, as the terrain is relatively flat and the compression has little impact on elevation cues. In steep, densely forested areas (e.g., local areas in OpenGF Test III), the method maintains high $OA$ by keeping the central region intact. However, the compression‑induced misleading contextual cues from the periphery, together with the limited effectiveness of the HAG loss under extremely sparse ground conditions, degrade $RMSE$.

\subsection{Limitations and future directions}
\label{sec:limitations}
Despite the demonstrated strengths of the proposed three core components, each has specific limitations that point to valuable directions for future research.

\subsubsection{Terrain dependence and fixed parameters of context compression}
The effectiveness of the context compression module degrades as terrain steepness increases, as demonstrated by the performance gradient in Section~\ref{sec:opengfresult}. Moreover, the compression radii are fixed during training. While the method generalizes reasonably to different terrain types, its adaptability to scenes with substantially different spatial scales remains limited. Future work could explore adaptive or learnable compression parameters.

\subsubsection{Backbone capacity and module effectiveness in steep terrains}
The performance of SparseGF in steep forested areas is limited by two factors: the backbone's representational capacity and the limited benefit of the context compression module due to scarce large objects. As a result, performance gains rely primarily on the point-wise refinement module and the height-aware loss, which alone cannot fully compensate for the backbone's limitations. Future work could integrate the proposed modules with more powerful backbones to improve performance in challenging natural terrains while preserving leading results in complex urban scenes.

\subsubsection{Limitations of height-aware loss under sparse ground references}
The height-aware loss is effective only when sufficient ground references exist. In areas with extremely sparse ground points (e.g., OpenGF Test III), its effectiveness is severely limited. Future work could explore strategies that incorporate direct DSM-to-DTM mapping~\citep{amirkolaee2022dtm} to provide more robust elevation constraints.

\section{Conclusion}
\label{sec:conclusion}
To address the primary bottleneck of DL-based GF across diverse landscapes, this study introduces SparseGF, a height-aware sparse segmentation framework enhanced with context compression. The framework features three key innovations: (1) a convex-mirror-inspired context compression module that condenses expansive contexts while preserving central details; (2) a hybrid sparse voxel-point network that isolates the compressed periphery from classification and loss computation to mitigate geometric distortion; and (3) a height-aware loss function that enforces topographic elevation priors to suppress tall-object misclassification.

Evaluated on two large-scale public ALS datasets, SparseGF demonstrates robust performance across varied terrains. In complex urban scenes, it achieves leading results, substantially outperforming existing methods. On mixed terrains with moderate slopes and vegetation, it performs on par with the best existing methods. Even in densely forested, steep mountainous areas, where context compression faces the greatest challenge, the framework maintains a reasonable accuracy, underscoring its overall robustness.

This study provides a methodological advancement in DL-based GF, enabling reliable DTM generation for large-scale environmental monitoring. Future work may incorporate DTM regression, develop adaptive compression mechanisms, and employ more powerful backbones. Another promising direction is to extend the proposed framework to panoptic segmentation of ALS point clouds, which would enable comprehensive scene understanding for applications such as urban environmental monitoring, infrastructure planning, and digital twin modeling.

%\end{linenumbers}

\section*{Acknowledgments}
This work was supported in part by the National Natural Science Foundation of China (Grant No. 42471477). The authors acknowledge the use of an AI-based language tool (DeepSeek) for grammatical refinement and expression improvement. All content has been reviewed and is the sole responsibility of the authors.

\appendix
\section{Formula derivation for context compression}
\label{sec:formula}

\begin{figure}[ht]
    \setlength{\abovecaptionskip}{0.cm}
    \setlength{\belowcaptionskip}{-0.cm}
    \centering
    \includegraphics[width=0.8\linewidth]{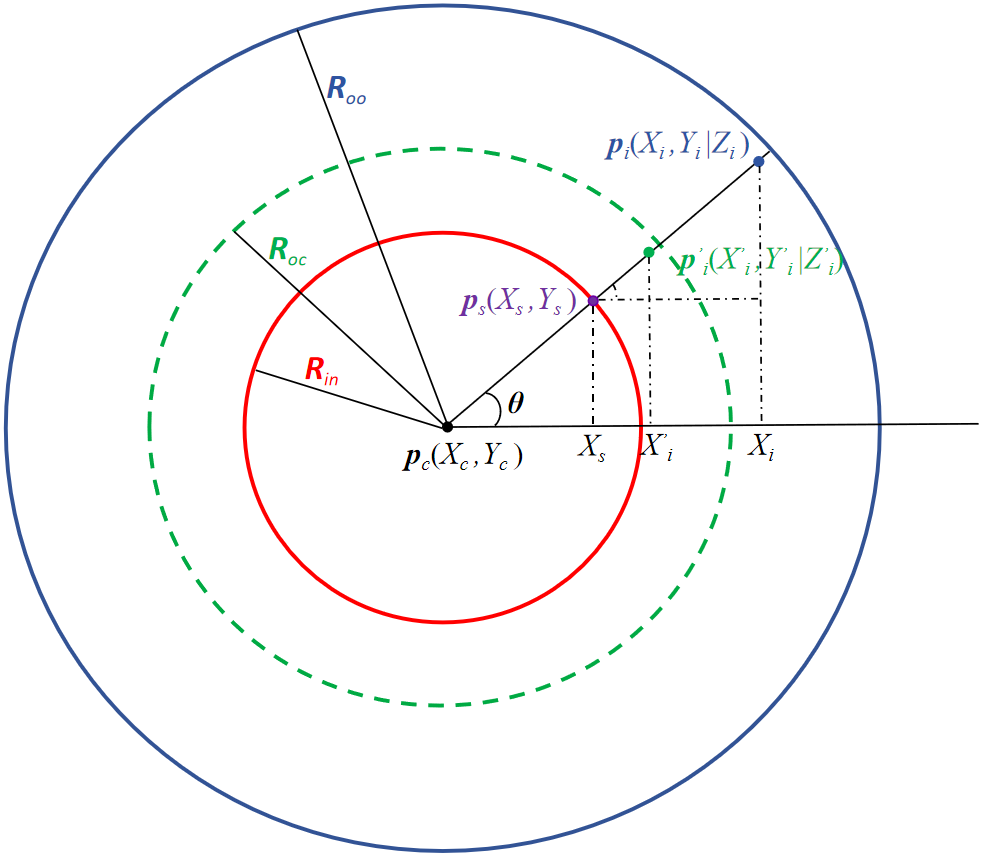}
    \caption{A schematic of the formula derivation for context compression.}
    \label{fig:sc_f}
\end{figure} 

To formalize the context compression transformation, we derive the mapping from an original point \( P_i(X_i, Y_i, Z_i) \) in the compression zone to its compressed counterpart \( P'_i(X'_i, Y'_i, Z'_i) \). As illustrated in Fig.~\ref{fig:sc_f}, the key geometric parameters are defined as follows: \( R_{oo} \), \( R_{oc} \), and \( R_{in} \) are the radii of the outer original, outer compressed, and inner invariant boundaries, with \( p_c(X_c, Y_c) \) being the center. For the XY-plane projections \( p_i \) of \( P_i \) and \( p'_i \) of \( P'_i \), define \( p_s(X_s, Y_s) \) as the intersection of segment \( \overline{p_i p_c} \) with the inner boundary, and \( \theta \) as the angle between this segment and the horizontal axis. Their distances from \( p_s \) are denoted as \( D_{si} \) and \( D'_{si} \), respectively. The coordinate transformation is then derived geometrically as follows (Eqs.~\ref{eq:e1}–\ref{eq:e10}).

Based on the geometric similarity shown in Fig.~\ref{fig:sc_f}, we derive:
\begin{equation} \label{eq:e1}
\frac{X'_i - X_s}{X_i - X_s} = \frac{D'_{si}}{D_{si}}
\end{equation}

Since the compression is applied linearly along the radial direction, the following relationship holds:
\begin{equation} \label{eq:e2}
\frac{D'_{si}}{R_{oc} - R_{in}} = \frac{D_{si}}{R_{oo} - R_{in}}
\end{equation}

Substituting Eq. \ref{eq:e2} into Eq. \ref{eq:e1} yields:
\begin{equation} \label{eq:e3}
\frac{X'_i - X_s}{X_i - X_s} = \frac{R_{oc} - R_{in}}{R_{oo} - R_{in}}
\end{equation}

Rewriting Eq. \ref{eq:e3}, we obtain:
\begin{equation} \label{eq:e4}
X'_i = \frac{R_{oc} - R_{in}}{R_{oo} - R_{in}} \cdot X_i + (1-\frac{R_{oc} - R_{in}}{R_{oo} - R_{in}}) \cdot X_s
\end{equation}

From the geometric relationships in Fig.~\ref{fig:sc_f}, we have:
\begin{equation} \label{eq:e5}
X_s = R_{in} \cdot \cos\theta + X_c
\end{equation}
\begin{equation} \label{eq:e6}
\cos\theta = \frac{X_i - X_c}{D_{ci}}
\end{equation}

Here, \( D_{ci} \) is the distance between \( p_i \) and \( p_c \), given by:
\begin{equation} \label{eq:e7}
D_{ci} = \sqrt{(X_i - X_c)^2 + (Y_i - Y_c)^2}
\end{equation}

Substituting Eqs. \ref{eq:e5}–\ref{eq:e7} into Eq. \ref{eq:e4}, the compressed X-coordinate is expressed as:
\begin{equation} \label{eq:e8}
X'_i = \frac{R_{oc} - R_{in}}{R_{oo} - R_{in}} \cdot X_i + \left(1 - \frac{R_{oc} - R_{in}}{R_{oo} - R_{in}}\right) \cdot \left( R_{in} \cdot \frac{X_i - X_c}{\sqrt{(X_i - X_c)^2 + (Y_i - Y_c)^2}} + X_c \right)
\end{equation}

Similarly, the Y-coordinate after compression is given by:
\begin{equation} \label{eq:e9}
Y'_i = \frac{R_{oc} - R_{in}}{R_{oo} - R_{in}} \cdot Y_i + \left(1 - \frac{R_{oc} - R_{in}}{R_{oo} - R_{in}}\right) \cdot \left( R_{in} \cdot \frac{Y_i - Y_c}{\sqrt{(X_i - X_c)^2 + (Y_i - Y_c)^2}} + Y_c \right)
\end{equation}

Importantly, to preserve the discriminability between ground and non-ground points, the elevation coordinate remains unchanged:
\begin{equation} \label{eq:e10}
Z'_i = Z_i
\end{equation}

\section{Hyperparameter sensitivity analysis}
\label{sec:hyperparameter}

We provide a detailed sensitivity analysis of three key hyperparameters on the OpenGF dataset, namely the balancing coefficient \(\lambda\) in the combined loss \(L_{\text{total}} = (1-\lambda)L_{\text{cls}} + \lambda L_{\text{hag}}\), the number of HAG bins, and the outer compressed radius \(R_{oc}\). Figure~\ref{fig:hyper_sensitivity} summarizes the $OA$ on Tests I, II, and III for each configuration.

For non‑default hyperparameter settings, we fine‑tune the model from the default pre‑trained checkpoint (trained with the default hyperparameters) using a reduced number of epochs to isolate the effect of the changed hyperparameter. Specifically, for \(\lambda\) and the number of HAG bins, we use a fixed learning rate of \(1 \times 10^{-3}\) for 3 epochs; for \(R_{oc}\), we use \(1 \times 10^{-4}\) for 5 epochs. This ensures that all comparisons share the same initial model and that observed differences are not due to varying training length.

\begin{figure}[H]
\centering
\includegraphics[width=\linewidth]{Figures/hyperparas.png}
\caption{Sensitivity analysis of key hyperparameters: (a) balancing coefficient \(\lambda\), (b) number of HAG bins, (c)  outer compressed radius \(R_{oc}\).}
\label{fig:hyper_sensitivity}
\end{figure}

\textbf{Balancing coefficient \(\lambda\).}
We test \(\lambda = 0.25\), 0.5 (the default), and 0.75. The $OA$ varies by less than 0.5\% on average across the three test sites, indicating that the method is stable with respect to \(\lambda\) in the range \([0.25, 0.75]\). The default \(\lambda = 0.5\) is therefore a reasonable choice.

\textbf{Number of HAG bins.}
We test 4, 6 (the default), and 8 bins while keeping the bin boundaries proportionally scaled to the height ranges defined in the implementation. The $OA$ varies by less than 0.6\% on average across the three test sites, indicating that the discretization is robust to the choice of bin number. The default 6‑bin setting is therefore a reasonable choice.

\textbf{Outer compressed radius \(R_{oc}\).}
We test \(R_{oc} = 40\) m, 44 m (the default), and 50 m. The $OA$ varies by less than 0.4\% on average across the three test sites, indicating that the method is stable with respect to \(R_{oc}\) in the tested range. The default \(R_{oc} = 44\) is therefore a reasonable choice.

Overall, the method exhibits robust and stable performance across all tested hyperparameters. The default settings (\(\lambda=0.5\), 6 HAG bins, \(R_{oc}=44\) m) are well justified.

\section*{Data availability}
The datasets used in this study are available upon request. The core network code of SparseGF and the pre-trained models will be made publicly available at \url{https://github.com/Nathan-UW/SparseGF} upon publication. The complete pipeline can be reproduced using the provided code and the datasets.

\bibliographystyle{elsarticle-harv} 
\bibliography{cas-refs}

\begin{thebibliography}{51}
\expandafter\ifx\csname natexlab\endcsname\relax\def\natexlab#1{#1}\fi
\providecommand{\url}[1]{\texttt{#1}}
\providecommand{\href}[2]{#2}
\providecommand{\path}[1]{#1}
\providecommand{\DOIprefix}{doi:}
\providecommand{\ArXivprefix}{arXiv:}
\providecommand{\URLprefix}{URL: }
\providecommand{\Pubmedprefix}{pmid:}
\providecommand{\doi}[1]{\href{http://dx.doi.org/#1}{\path{#1}}}
\providecommand{\Pubmed}[1]{\href{pmid:#1}{\path{#1}}}
\providecommand{\bibinfo}[2]{#2}
\ifx\xfnm\relax \def\xfnm[#1]{\unskip,\space#1}\fi
%Type = Article
\bibitem[{{Amini Amirkolaee} et~al.(2022){Amini Amirkolaee}, Arefi, Ahmadlou and Raikwar}]{amirkolaee2022dtm}
\bibinfo{author}{{Amini Amirkolaee}, H.}, \bibinfo{author}{Arefi, H.}, \bibinfo{author}{Ahmadlou, M.}, \bibinfo{author}{Raikwar, V.}, \bibinfo{year}{2022}.
\newblock \bibinfo{title}{{DTM} extraction from {DSM} using a multi-scale {DTM} fusion strategy based on deep learning}.
\newblock \bibinfo{journal}{Remote Sens. Environ.} \bibinfo{volume}{274}, \bibinfo{pages}{113014}.
%Type = Article
\bibitem[{Arief et~al.(2019)Arief, Indahl, Strand and Tveite}]{ARIEF2019a}
\bibinfo{author}{Arief, H.A.}, \bibinfo{author}{Indahl, U.G.}, \bibinfo{author}{Strand, G.H.}, \bibinfo{author}{Tveite, H.}, \bibinfo{year}{2019}.
\newblock \bibinfo{title}{Addressing overfitting on point cloud classification using atrous xcrf}.
\newblock \bibinfo{journal}{ISPRS J. Photogramm. Remote Sens.} \bibinfo{volume}{155}, \bibinfo{pages}{90--101}.
%Type = Article
\bibitem[{Axelsson(2000)}]{axelsson2000generation}
\bibinfo{author}{Axelsson, P.}, \bibinfo{year}{2000}.
\newblock \bibinfo{title}{{DEM} generation from laser scanner data using adaptive {TIN} models}.
\newblock \bibinfo{journal}{Int. Arch. Photogramm. Remote Sens.} \bibinfo{volume}{XXXIII, Part B4}, \bibinfo{pages}{110--117}.
%Type = Inproceedings
\bibitem[{Bartels et~al.(2006)Bartels, Wei and Mason}]{bartels2006dtm}
\bibinfo{author}{Bartels, M.}, \bibinfo{author}{Wei, H.}, \bibinfo{author}{Mason, D.C.}, \bibinfo{year}{2006}.
\newblock \bibinfo{title}{{DTM} generation from {LiDAR} data using skewness balancing}, in: \bibinfo{booktitle}{Proc. Int. Conf. on Pattern Recog.}, pp. \bibinfo{pages}{566--569}.
%Type = Inproceedings
\bibitem[{Berman et~al.(2018)Berman, Triki and Blaschko}]{Berman2018The}
\bibinfo{author}{Berman, M.}, \bibinfo{author}{Triki, A.R.}, \bibinfo{author}{Blaschko, M.B.}, \bibinfo{year}{2018}.
\newblock \bibinfo{title}{The lovasz-softmax loss: A tractable surrogate for the optimization of the intersection-over-union measure in neural networks}, in: \bibinfo{booktitle}{Proc. IEEE Conf. Comput. Vis. Pattern Recog.}, pp. \bibinfo{pages}{4413--4421}.
%Type = Article
\bibitem[{Beumier and Idrissa(2016)}]{Beumier2016Digital}
\bibinfo{author}{Beumier, C.}, \bibinfo{author}{Idrissa, M.}, \bibinfo{year}{2016}.
\newblock \bibinfo{title}{Digital terrain models derived from digital surface model uniform regions in urban areas}.
\newblock \bibinfo{journal}{Int. J. Remote Sens.} \bibinfo{volume}{37}, \bibinfo{pages}{3477--3493}.
%Type = Article
\bibitem[{Bigdeli et~al.(2018)Bigdeli, Amirkolaee and Pahlavani}]{Bigdeli2018DTM}
\bibinfo{author}{Bigdeli, B.}, \bibinfo{author}{Amirkolaee, H.A.}, \bibinfo{author}{Pahlavani, P.}, \bibinfo{year}{2018}.
\newblock \bibinfo{title}{{DTM} extraction under forest canopy using {LiDAR} data and a modified invasive weed optimization algorithm}.
\newblock \bibinfo{journal}{Remote Sens. Environ.} \bibinfo{volume}{216}, \bibinfo{pages}{289--300}.
%Type = Article
\bibitem[{Chen et~al.(2014)Chen, Li, Wang, Chen and Liu}]{CHEN2014291}
\bibinfo{author}{Chen, W.}, \bibinfo{author}{Li, X.}, \bibinfo{author}{Wang, Y.}, \bibinfo{author}{Chen, G.}, \bibinfo{author}{Liu, S.}, \bibinfo{year}{2014}.
\newblock \bibinfo{title}{Forested landslide detection using {LiDAR} data and the random forest algorithm: {A case study of the Three Gorges, China}}.
\newblock \bibinfo{journal}{Remote Sens. Environ.} \bibinfo{volume}{152}, \bibinfo{pages}{291--301}.
%Type = Inproceedings
\bibitem[{Choy et~al.(2019)Choy, Gwak and Savarese}]{choy20194d}
\bibinfo{author}{Choy, C.}, \bibinfo{author}{Gwak, J.}, \bibinfo{author}{Savarese, S.}, \bibinfo{year}{2019}.
\newblock \bibinfo{title}{{4D} spatio-temporal convnets: Minkowski convolutional neural networks}, in: \bibinfo{booktitle}{Proc. IEEE Conf. Comput. Vis. Pattern Recog.}, pp. \bibinfo{pages}{3075--3084}.
%Type = Article
\bibitem[{Dai et~al.(2023)Dai, Hu, Shu, Qin and Zhang}]{Dai2023deep}
\bibinfo{author}{Dai, H.}, \bibinfo{author}{Hu, X.}, \bibinfo{author}{Shu, Z.}, \bibinfo{author}{Qin, N.}, \bibinfo{author}{Zhang, J.}, \bibinfo{year}{2023}.
\newblock \bibinfo{title}{Deep ground filtering of large-scale {ALS} point clouds via iterative sequential ground prediction}.
\newblock \bibinfo{journal}{Remote Sens.} \bibinfo{volume}{15}, \bibinfo{pages}{961}.
%Type = Article
\bibitem[{Dai et~al.(2024)Dai, Hu, Zhang, Shu, Xu and Du}]{Dai2024large}
\bibinfo{author}{Dai, H.}, \bibinfo{author}{Hu, X.}, \bibinfo{author}{Zhang, J.}, \bibinfo{author}{Shu, Z.}, \bibinfo{author}{Xu, J.}, \bibinfo{author}{Du, J.}, \bibinfo{year}{2024}.
\newblock \bibinfo{title}{Large-scale als point cloud segmentation via projection-based context embedding}.
\newblock \bibinfo{journal}{IEEE Trans. Geosci. Remote Sens.} \bibinfo{volume}{62}, \bibinfo{pages}{1--16}.
%Type = Article
\bibitem[{Evans and Hudak(2007)}]{Evans2007multiscale}
\bibinfo{author}{Evans, J.S.}, \bibinfo{author}{Hudak, A.T.}, \bibinfo{year}{2007}.
\newblock \bibinfo{title}{A multiscale curvature algorithm for classifying discrete return {LiDAR} in forested environments}.
\newblock \bibinfo{journal}{IEEE Trans. Geosci. Remote Sens.} \bibinfo{volume}{45}, \bibinfo{pages}{1029--1038}.
%Type = Inproceedings
\bibitem[{Fan et~al.(2021)Fan, Dong, Zhu, Lv, Ye and Wang}]{fan2021scf}
\bibinfo{author}{Fan, S.}, \bibinfo{author}{Dong, Q.}, \bibinfo{author}{Zhu, F.}, \bibinfo{author}{Lv, Y.}, \bibinfo{author}{Ye, P.}, \bibinfo{author}{Wang, F.Y.}, \bibinfo{year}{2021}.
\newblock \bibinfo{title}{{SCF-Net}: Learning spatial contextual features for large-scale point cloud segmentation}, in: \bibinfo{booktitle}{Proc. IEEE Conf. Comput. Vis. Pattern Recog.}, pp. \bibinfo{pages}{14504--14513}.
%Type = Article
\bibitem[{Fareed et~al.(2023)Fareed, Flores and Das}]{Fareed2023Analysis}
\bibinfo{author}{Fareed, N.}, \bibinfo{author}{Flores, J.P.}, \bibinfo{author}{Das, A.K.}, \bibinfo{year}{2023}.
\newblock \bibinfo{title}{Analysis of {UAS-LiDAR} ground points classification in agricultural fields using traditional algorithms and {PointCNN}}.
\newblock \bibinfo{journal}{Remote Sens.} \bibinfo{volume}{15}, \bibinfo{pages}{483}.
%Type = Article
\bibitem[{Gevaert et~al.(2018)Gevaert, Persello, Nex and Vosselman}]{gevaert2018deep}
\bibinfo{author}{Gevaert, C.}, \bibinfo{author}{Persello, C.}, \bibinfo{author}{Nex, F.}, \bibinfo{author}{Vosselman, G.}, \bibinfo{year}{2018}.
\newblock \bibinfo{title}{A deep learning approach to {DTM} extraction from imagery using rule-based training labels}.
\newblock \bibinfo{journal}{ISPRS J. Photogramm. Remote Sens.} \bibinfo{volume}{142}, \bibinfo{pages}{106--123}.
%Type = Inproceedings
\bibitem[{Girshick(2015)}]{Girshick2015f}
\bibinfo{author}{Girshick, R.}, \bibinfo{year}{2015}.
\newblock \bibinfo{title}{{Fast R-CNN}}, in: \bibinfo{booktitle}{Proc. IEEE Int. Conf. Comput. Vis.}, pp. \bibinfo{pages}{1440--1448}.
%Type = Article
\bibitem[{Hingee et~al.(2016)Hingee, Caccetta, Caccetta, Wu and Devereaux}]{Hingee2016DIGITAL}
\bibinfo{author}{Hingee, K.}, \bibinfo{author}{Caccetta, P.}, \bibinfo{author}{Caccetta, L.}, \bibinfo{author}{Wu, X.}, \bibinfo{author}{Devereaux, D.}, \bibinfo{year}{2016}.
\newblock \bibinfo{title}{Digital terrain from a two-step segmentation and outlier-based algorithm}.
\newblock \bibinfo{journal}{Int. Arch. Photogramm. Remote Sens. Spatial Inf. Sci.} \bibinfo{volume}{XLI-B3}, \bibinfo{pages}{233--239}.
%Type = Inproceedings
\bibitem[{Hu et~al.(2020)Hu, Yang, Xie, Rosa, Guo, Wang, Trigoni and Markham}]{hu2020randla}
\bibinfo{author}{Hu, Q.}, \bibinfo{author}{Yang, B.}, \bibinfo{author}{Xie, L.}, \bibinfo{author}{Rosa, S.}, \bibinfo{author}{Guo, Y.}, \bibinfo{author}{Wang, Z.}, \bibinfo{author}{Trigoni, N.}, \bibinfo{author}{Markham, A.}, \bibinfo{year}{2020}.
\newblock \bibinfo{title}{{RandLA-Net}: Efficient semantic segmentation of large-scale point clouds}, in: \bibinfo{booktitle}{Proc. IEEE Conf. Comput. Vis. Pattern Recog.}, pp. \bibinfo{pages}{11105--11114}.
%Type = Article
\bibitem[{Hu and Yuan(2016)}]{hu2016deep}
\bibinfo{author}{Hu, X.}, \bibinfo{author}{Yuan, Y.}, \bibinfo{year}{2016}.
\newblock \bibinfo{title}{Deep-learning-based classification for {DTM} extraction from {ALS} point cloud}.
\newblock \bibinfo{journal}{Remote sens.} \bibinfo{volume}{8}, \bibinfo{pages}{730}.
%Type = Article
\bibitem[{Jahromi et~al.(2011)Jahromi, Zoej, Mohammadzadeh and Sadeghian}]{jahromi2011novel}
\bibinfo{author}{Jahromi, A.B.}, \bibinfo{author}{Zoej, M.J.V.}, \bibinfo{author}{Mohammadzadeh, A.}, \bibinfo{author}{Sadeghian, S.}, \bibinfo{year}{2011}.
\newblock \bibinfo{title}{A novel filtering algorithm for bare-earth extraction from airborne laser scanning data using an artificial neural network}.
\newblock \bibinfo{journal}{IEEE J. Sel. Top. Appl. Earth Obs. Remote Sens.} \bibinfo{volume}{4}, \bibinfo{pages}{836--843}.
%Type = Article
\bibitem[{Jin et~al.(2020)Jin, Su, Zhao, Hu and Guo}]{jin2020point}
\bibinfo{author}{Jin, S.}, \bibinfo{author}{Su, Y.}, \bibinfo{author}{Zhao, X.}, \bibinfo{author}{Hu, T.}, \bibinfo{author}{Guo, Q.}, \bibinfo{year}{2020}.
\newblock \bibinfo{title}{A point-based fully convolutional neural network for airborne {LiDAR} ground point filtering in forested environments}.
\newblock \bibinfo{journal}{IEEE J. Sel. Top. Appl. Earth Obs. Remote Sens.} \bibinfo{volume}{13}, \bibinfo{pages}{3958--3974}.
%Type = Article
\bibitem[{Kilian et~al.(1996)Kilian, Haala, Englich and Iii}]{Kilian1996capture}
\bibinfo{author}{Kilian, J.}, \bibinfo{author}{Haala, N.}, \bibinfo{author}{Englich, M.}, \bibinfo{author}{Iii, C.}, \bibinfo{year}{1996}.
\newblock \bibinfo{title}{Capture and evaluation of airborne laser scanner data}.
\newblock \bibinfo{journal}{Int. Arch. Photogramm. Remote Sens.} \bibinfo{volume}{XXXI}, \bibinfo{pages}{383--388}.
%Type = Article
\bibitem[{Le~Saux et~al.(2019)Le~Saux, Yokoya, Haensch and Brown}]{Saux2019Gr}
\bibinfo{author}{Le~Saux, B.}, \bibinfo{author}{Yokoya, N.}, \bibinfo{author}{Haensch, R.}, \bibinfo{author}{Brown, M.}, \bibinfo{year}{2019}.
\newblock \bibinfo{title}{{2019 IEEE GRSS Data Fusion Contest}: Large-scale semantic 3d reconstruction [technical committees]}.
\newblock \bibinfo{journal}{IEEE Geosci. Remote Sens. Mag.} \bibinfo{volume}{7}, \bibinfo{pages}{33--36}.
%Type = Article
\bibitem[{Li et~al.(2025)Li, Tang, Ma, Zhu, Gong, Ruhaiyem and Li}]{Li2025p}
\bibinfo{author}{Li, J.}, \bibinfo{author}{Tang, F.}, \bibinfo{author}{Ma, L.}, \bibinfo{author}{Zhu, C.}, \bibinfo{author}{Gong, Z.}, \bibinfo{author}{Ruhaiyem, N.I.R.}, \bibinfo{author}{Li, J.}, \bibinfo{year}{2025}.
\newblock \bibinfo{title}{Point-sct: A multiscale spatial convolution-swin transformer network for point cloud ground filtering in complex mountainous terrains}.
\newblock \bibinfo{journal}{IEEE Trans. Geosci. Remote Sens.} \bibinfo{volume}{63}, \bibinfo{pages}{1--18}.
%Type = Article
\bibitem[{Lu et~al.(2009)Lu, Murphy, Little, Sheffer and Fu}]{lu2009hybrid}
\bibinfo{author}{Lu, W.L.}, \bibinfo{author}{Murphy, K.P.}, \bibinfo{author}{Little, J.J.}, \bibinfo{author}{Sheffer, A.}, \bibinfo{author}{Fu, H.}, \bibinfo{year}{2009}.
\newblock \bibinfo{title}{A hybrid conditional random field for estimating the underlying ground surface from airborne {LiDAR} data}.
\newblock \bibinfo{journal}{IEEE Trans. Geosci. Remote Sens.} \bibinfo{volume}{47}, \bibinfo{pages}{2913--2922}.
%Type = Article
\bibitem[{McCarley et~al.(2020)McCarley, Hudak, Sparks, Vaillant, Meddens, Trader, Mauro, Kreitler and Boschetti}]{mccarley2020estimating}
\bibinfo{author}{McCarley, T.R.}, \bibinfo{author}{Hudak, A.T.}, \bibinfo{author}{Sparks, A.M.}, \bibinfo{author}{Vaillant, N.M.}, \bibinfo{author}{Meddens, A.J.}, \bibinfo{author}{Trader, L.}, \bibinfo{author}{Mauro, F.}, \bibinfo{author}{Kreitler, J.}, \bibinfo{author}{Boschetti, L.}, \bibinfo{year}{2020}.
\newblock \bibinfo{title}{Estimating wildfire fuel consumption with multitemporal airborne laser scanning data and demonstrating linkage with {MODIS}-derived fire radiative energy}.
\newblock \bibinfo{journal}{Remote Sens. Environ.} \bibinfo{volume}{251}, \bibinfo{pages}{112114}.
%Type = Article
\bibitem[{Mihu-Pintilie et~al.(2025)Mihu-Pintilie, Urzică, Stoleriu and Pricop}]{Mihu2025Int}
\bibinfo{author}{Mihu-Pintilie, A.}, \bibinfo{author}{Urzică, A.}, \bibinfo{author}{Stoleriu, C.C.}, \bibinfo{author}{Pricop, C.I.}, \bibinfo{year}{2025}.
\newblock \bibinfo{title}{Integrating lidar-derived dem, rainfall radar data, and sar imagery for 2d hec-ras modelling to assess the severity of pluvial flash floods induced by storm boris in se romania}.
\newblock \bibinfo{journal}{Geomat. Nat. Haz. RISK} \bibinfo{volume}{16}, \bibinfo{pages}{2488190}.
%Type = Article
\bibitem[{{\"O}zcan and {\"U}nsalan(2016)}]{ozcan2016lidar}
\bibinfo{author}{{\"O}zcan, A.H.}, \bibinfo{author}{{\"U}nsalan, C.}, \bibinfo{year}{2016}.
\newblock \bibinfo{title}{{LiDAR} data filtering and {DTM} generation using empirical mode decomposition}.
\newblock \bibinfo{journal}{IEEE J. Sel. Top. Appl. Earth Obs. Remote Sens.} \bibinfo{volume}{10}, \bibinfo{pages}{360--371}.
%Type = Article
\bibitem[{Pfeifer et~al.(1999)Pfeifer, Reiter, Briese and Rieger}]{Pfeifer1999Interpolation}
\bibinfo{author}{Pfeifer, N.}, \bibinfo{author}{Reiter, T.}, \bibinfo{author}{Briese, C.}, \bibinfo{author}{Rieger, W.}, \bibinfo{year}{1999}.
\newblock \bibinfo{title}{Interpolation of high quality ground models from laser scanner data in forested areas}.
\newblock \bibinfo{journal}{Int. Arch. Photogramm. Remote Sens.} \bibinfo{volume}{32}, \bibinfo{pages}{31–36}.
%Type = Article
\bibitem[{Pingel et~al.(2013)Pingel, Clarke and McBride}]{pingel2013improved}
\bibinfo{author}{Pingel, T.J.}, \bibinfo{author}{Clarke, K.C.}, \bibinfo{author}{McBride, W.A.}, \bibinfo{year}{2013}.
\newblock \bibinfo{title}{An improved simple morphological filter for the terrain classification of airborne {LIDAR} data}.
\newblock \bibinfo{journal}{ISPRS J. Photogramm. Remote Sens.} \bibinfo{volume}{77}, \bibinfo{pages}{21--30}.
%Type = Article
\bibitem[{Qi et~al.(2017)Qi, Yi, Su and Guibas}]{qi2017pointnet++}
\bibinfo{author}{Qi, C.R.}, \bibinfo{author}{Yi, L.}, \bibinfo{author}{Su, H.}, \bibinfo{author}{Guibas, L.J.}, \bibinfo{year}{2017}.
\newblock \bibinfo{title}{Point{N}et++: Deep hierarchical feature learning on point sets in a metric space}.
\newblock \bibinfo{journal}{Proc. Adv. Neural Inf. Process. Syst.} \bibinfo{volume}{30}, \bibinfo{pages}{5105–5114}.
%Type = Article
\bibitem[{Qin et~al.(2019)Qin, Hu, Wang, Shan and Li}]{Qin2019Semantic}
\bibinfo{author}{Qin, N.}, \bibinfo{author}{Hu, X.}, \bibinfo{author}{Wang, P.}, \bibinfo{author}{Shan, J.}, \bibinfo{author}{Li, Y.}, \bibinfo{year}{2019}.
\newblock \bibinfo{title}{Semantic labeling of als point cloud via learning voxel and pixel representations}.
\newblock \bibinfo{journal}{IEEE Geosci. Remote Sens. Lett.} \bibinfo{volume}{PP}, \bibinfo{pages}{1--5}.
%Type = Article
\bibitem[{Qin et~al.(2023a)Qin, Tan, Guan, Wang, Ma, Tao, Fatholahi, Hu and Li}]{QIN2023tow}
\bibinfo{author}{Qin, N.}, \bibinfo{author}{Tan, W.}, \bibinfo{author}{Guan, H.}, \bibinfo{author}{Wang, L.}, \bibinfo{author}{Ma, L.}, \bibinfo{author}{Tao, P.}, \bibinfo{author}{Fatholahi, S.}, \bibinfo{author}{Hu, X.}, \bibinfo{author}{Li, J.}, \bibinfo{year}{2023}a.
\newblock \bibinfo{title}{Towards intelligent ground filtering of large-scale topographic point clouds: A comprehensive survey}.
\newblock \bibinfo{journal}{Int. J. Appl. Earth Obs. Geoinf.} \bibinfo{volume}{125}, \bibinfo{pages}{103566}.
%Type = Article
\bibitem[{Qin et~al.(2023b)Qin, Tan, Ma, Zhang, Guan and Li}]{qin2023deep}
\bibinfo{author}{Qin, N.}, \bibinfo{author}{Tan, W.}, \bibinfo{author}{Ma, L.}, \bibinfo{author}{Zhang, D.}, \bibinfo{author}{Guan, H.}, \bibinfo{author}{Li, J.}, \bibinfo{year}{2023}b.
\newblock \bibinfo{title}{Deep learning for filtering the ground from {ALS} point clouds: A dataset, evaluations and issues}.
\newblock \bibinfo{journal}{ISPRS J. Photogramm. Remote Sens.} \bibinfo{volume}{202}, \bibinfo{pages}{246--261}.
%Type = Inproceedings
\bibitem[{Qin et~al.(2021)Qin, Tan, Ma, Zhang and Li}]{qin2021opengf}
\bibinfo{author}{Qin, N.}, \bibinfo{author}{Tan, W.}, \bibinfo{author}{Ma, L.}, \bibinfo{author}{Zhang, D.}, \bibinfo{author}{Li, J.}, \bibinfo{year}{2021}.
\newblock \bibinfo{title}{{OpenGF}: An ultra-large-scale ground filtering dataset built upon open {ALS} point clouds around the world}, in: \bibinfo{booktitle}{Proc. IEEE Conf. Comput. Vis. Pattern Recog. Workshops}, pp. \bibinfo{pages}{1082--1091}.
%Type = Article
\bibitem[{Razak et~al.(2011)Razak, Straatsma, {van Westen}, Malet and {de Jong}}]{RAZAK2011186}
\bibinfo{author}{Razak, K.}, \bibinfo{author}{Straatsma, M.}, \bibinfo{author}{{van Westen}, C.}, \bibinfo{author}{Malet, J.P.}, \bibinfo{author}{{de Jong}, S.}, \bibinfo{year}{2011}.
\newblock \bibinfo{title}{Airborne laser scanning of forested landslides characterization: Terrain model quality and visualization}.
\newblock \bibinfo{journal}{Geomorphology} \bibinfo{volume}{126}, \bibinfo{pages}{186--200}.
%Type = Article
\bibitem[{Rizaldy et~al.(2018)Rizaldy, Persello, Gevaert, Oude~Elberink and Vosselman}]{rizaldy2018fully}
\bibinfo{author}{Rizaldy, A.}, \bibinfo{author}{Persello, C.}, \bibinfo{author}{Gevaert, C.}, \bibinfo{author}{Oude~Elberink, S.}, \bibinfo{author}{Vosselman, G.}, \bibinfo{year}{2018}.
\newblock \bibinfo{title}{Ground and multi-class classification of airborne laser scanner point clouds using fully convolutional networks}.
\newblock \bibinfo{journal}{Remote Sens.} \bibinfo{volume}{10}, \bibinfo{pages}{1723}.
%Type = Article
\bibitem[{Sithole(2001)}]{sithole2001filtering}
\bibinfo{author}{Sithole, G.}, \bibinfo{year}{2001}.
\newblock \bibinfo{title}{Filtering of laser altimetry data using a slope adaptive filter}.
\newblock \bibinfo{journal}{Int. Arch. Photogramm. Remote Sens.} \bibinfo{volume}{XXXIV-3/W4}, \bibinfo{pages}{203--210}.
%Type = Inproceedings
\bibitem[{Sithole and Vosselman(2005)}]{sithole_filtering_2005}
\bibinfo{author}{Sithole, G.}, \bibinfo{author}{Vosselman, G.}, \bibinfo{year}{2005}.
\newblock \bibinfo{title}{Filtering of airborne laser scanner data based on segmented point clouds}, in: \bibinfo{booktitle}{ISPRS WG III/3, III/4, V/3 Workshop ``Laserscanning 2005"}, pp. \bibinfo{pages}{66--71}.
%Type = Article
\bibitem[{Su et~al.(2015)Su, Sun, Zhong, Huang, Li, Zhu, Zhang, Wu and Zhu}]{Su2015hierarchical}
\bibinfo{author}{Su, W.}, \bibinfo{author}{Sun, Z.}, \bibinfo{author}{Zhong, R.}, \bibinfo{author}{Huang, J.}, \bibinfo{author}{Li, M.}, \bibinfo{author}{Zhu, J.}, \bibinfo{author}{Zhang, K.}, \bibinfo{author}{Wu, H.}, \bibinfo{author}{Zhu, D.}, \bibinfo{year}{2015}.
\newblock \bibinfo{title}{A new hierarchical moving curve-fitting algorithm for filtering lidar data for automatic {DTM} generation}.
\newblock \bibinfo{journal}{Int. J. Remote Sens.} \bibinfo{volume}{36}, \bibinfo{pages}{3616--3635}.
%Type = Article
\bibitem[{Susaki(2012)}]{susaki2012adaptive}
\bibinfo{author}{Susaki, J.}, \bibinfo{year}{2012}.
\newblock \bibinfo{title}{Adaptive slope filtering of airborne {LiDAR} data in urban areas for digital terrain model ({DTM}) generation}.
\newblock \bibinfo{journal}{Remote Sens.} \bibinfo{volume}{4}, \bibinfo{pages}{1804--1819}.
%Type = Article
\bibitem[{{Tan} et~al.(2024){Tan}, {Qin}, {Zhang}, {McGrath}, {Fortin} and {Li}}]{Tan2024r}
\bibinfo{author}{{Tan}, W.}, \bibinfo{author}{{Qin}, N.}, \bibinfo{author}{{Zhang}, Y.}, \bibinfo{author}{{McGrath}, H.}, \bibinfo{author}{{Fortin}, M.}, \bibinfo{author}{{Li}, J.}, \bibinfo{year}{2024}.
\newblock \bibinfo{title}{{A rapid high-resolution multi-sensory urban flood mapping framework via DEM upscaling}}.
\newblock \bibinfo{journal}{Remote Sens. Environ.} \bibinfo{volume}{301}, \bibinfo{pages}{113956}.
%Type = Inproceedings
\bibitem[{Thomas et~al.(2019)Thomas, Qi, Deschaud, Marcotegui, Goulette and Guibas}]{thomas2019kpconv}
\bibinfo{author}{Thomas, H.}, \bibinfo{author}{Qi, C.R.}, \bibinfo{author}{Deschaud, J.E.}, \bibinfo{author}{Marcotegui, B.}, \bibinfo{author}{Goulette, F.}, \bibinfo{author}{Guibas, L.J.}, \bibinfo{year}{2019}.
\newblock \bibinfo{title}{{KPConv}: Flexible and deformable convolution for point clouds}, in: \bibinfo{booktitle}{Proc. IEEE Int. Conf. Comput. Vis.}, pp. \bibinfo{pages}{6410--6419}.
%Type = Article
\bibitem[{Vosselman(2000)}]{vosselman2000slope}
\bibinfo{author}{Vosselman, G.}, \bibinfo{year}{2000}.
\newblock \bibinfo{title}{Slope based filtering of laser altimetry data}.
\newblock \bibinfo{journal}{Int. Arch. Photogramm. Remote Sens.} \bibinfo{volume}{XXXIII}, \bibinfo{pages}{935--942}.
%Type = Article
\bibitem[{Xiang et~al.(2024)Xiang, Wielgosz, Kontogianni, Peters, Puliti, Astrup and Schindler}]{XIANG2024Auto}
\bibinfo{author}{Xiang, B.}, \bibinfo{author}{Wielgosz, M.}, \bibinfo{author}{Kontogianni, T.}, \bibinfo{author}{Peters, T.}, \bibinfo{author}{Puliti, S.}, \bibinfo{author}{Astrup, R.}, \bibinfo{author}{Schindler, K.}, \bibinfo{year}{2024}.
\newblock \bibinfo{title}{Automated forest inventory: Analysis of high-density airborne lidar point clouds with {3D} deep learning}.
\newblock \bibinfo{journal}{Remote Sens. Environ.} \bibinfo{volume}{305}, \bibinfo{pages}{114078}.
%Type = Article
\bibitem[{Xu et~al.(2026)Xu, Dai, Su, Wang, Hu and Ke}]{Xu2026E}
\bibinfo{author}{Xu, J.}, \bibinfo{author}{Dai, H.}, \bibinfo{author}{Su, B.}, \bibinfo{author}{Wang, P.}, \bibinfo{author}{Hu, X.}, \bibinfo{author}{Ke, T.}, \bibinfo{year}{2026}.
\newblock \bibinfo{title}{End-to-end high-quality dem generation and ground filtering from als point clouds}.
\newblock \bibinfo{journal}{IEEE Transactions on Geoscience and Remote Sensing} \bibinfo{volume}{64}, \bibinfo{pages}{1--15}.
%Type = Article
\bibitem[{Yotsumata et~al.(2020)Yotsumata, Sakamoto and Satoh}]{Yotsumata2020Qua}
\bibinfo{author}{Yotsumata, T.}, \bibinfo{author}{Sakamoto, M.}, \bibinfo{author}{Satoh, T.}, \bibinfo{year}{2020}.
\newblock \bibinfo{title}{Quality improvement for airborne {LIDAR} data filtering based on deep learning method}.
\newblock \bibinfo{journal}{Int. Arch. Photogramm. Remote Sens. Spatial Inf. Sci.} \bibinfo{volume}{XLIII-B2-2020}, \bibinfo{pages}{355--360}.
%Type = Article
\bibitem[{Yousefhussien et~al.(2018)Yousefhussien, Kelbe, Ientilucci and Salvaggio}]{YOUSEFHUSSIEN2018a}
\bibinfo{author}{Yousefhussien, M.}, \bibinfo{author}{Kelbe, D.J.}, \bibinfo{author}{Ientilucci, E.J.}, \bibinfo{author}{Salvaggio, C.}, \bibinfo{year}{2018}.
\newblock \bibinfo{title}{A multi-scale fully convolutional network for semantic labeling of 3d point clouds}.
\newblock \bibinfo{journal}{ISPRS J. Photogramm. Remote Sens.} \bibinfo{volume}{143}, \bibinfo{pages}{191--204}.
%Type = Article
\bibitem[{Zhang et~al.(2003)Zhang, Chen, Whitman, Shyu, Yan and Zhang}]{zhang2003progressive}
\bibinfo{author}{Zhang, K.}, \bibinfo{author}{Chen, S.C.}, \bibinfo{author}{Whitman, D.}, \bibinfo{author}{Shyu, M.L.}, \bibinfo{author}{Yan, J.}, \bibinfo{author}{Zhang, C.}, \bibinfo{year}{2003}.
\newblock \bibinfo{title}{A progressive morphological filter for removing nonground measurements from airborne {LIDAR} data}.
\newblock \bibinfo{journal}{IEEE Trans. Geosci. Remote Sens.} \bibinfo{volume}{41}, \bibinfo{pages}{872--882}.
%Type = Article
\bibitem[{Zhang et~al.(2016)Zhang, Qi, Wan, Wang, Xie, Wang and Yan}]{zhang2016easy}
\bibinfo{author}{Zhang, W.}, \bibinfo{author}{Qi, J.}, \bibinfo{author}{Wan, P.}, \bibinfo{author}{Wang, H.}, \bibinfo{author}{Xie, D.}, \bibinfo{author}{Wang, X.}, \bibinfo{author}{Yan, G.}, \bibinfo{year}{2016}.
\newblock \bibinfo{title}{An easy-to-use airborne {LiDAR} data filtering method based on cloth simulation}.
\newblock \bibinfo{journal}{Remote Sens.} \bibinfo{volume}{8}, \bibinfo{pages}{501}.
%Type = Article
\bibitem[{Zhao et~al.(2016)Zhao, Guo, Su and Xue}]{ZHAO2016Improved}
\bibinfo{author}{Zhao, X.}, \bibinfo{author}{Guo, Q.}, \bibinfo{author}{Su, Y.}, \bibinfo{author}{Xue, B.}, \bibinfo{year}{2016}.
\newblock \bibinfo{title}{Improved progressive {TIN} densification filtering algorithm for airborne {LiDAR} data in forested areas}.
\newblock \bibinfo{journal}{ISPRS J. Photogramm. Remote Sens.} \bibinfo{volume}{117}, \bibinfo{pages}{79--91}.

\end{thebibliography}





% \end{thebibliography}
\end{document}